\documentclass{article}

\usepackage[final]{corl_2025} 

\usepackage{xspace}
\newcommand{\mindmap}{\textit{mindmap}\xspace}
\newcommand{\Mindmap}{\textit{Mindmap}\xspace}
\newcommand{\nvbloxtorch}{\textit{nvblox}\xspace}
\newcommand{\groot}{GR00T\xspace}

\newcommand{\cubestacking}{\textit{Cube Stacking}\xspace}
\newcommand{\mugindrawer}{\textit{Mug in Drawer}\xspace}
\newcommand{\drillinbox}{\textit{Drill in Box}\xspace}
\newcommand{\stickinbin}{\textit{Stick in Bin}\xspace}
\newcommand\rurl[1]{%
  \href{http://#1}{\nolinkurl{#1}}%
}

\usepackage[nolist,nohyperlinks]{acronym}
\usepackage{authblk}

\usepackage{xcolor}

\usepackage{hyperref}
\makeatletter
\newcommand{\Footnotemark}{%
  \footnotemark%
  \expandafter\global\expandafter\let\csname saved@Href@\thefootnote\endcsname\Hy@footnote@currentHref
}
\newcommand{\Footnotetext}[1]{%
  \expandafter\let\expandafter\Hy@footnote@currentHref\csname saved@Href@\thefootnote\endcsname%
  \footnotetext{#1}
}
\makeatother

\usepackage{enumitem}

\usepackage{amsmath}
\usepackage{amssymb}

\usepackage{minted}

\newcommand{\reffig}[1]{Fig.~\ref{#1}}

\newcommand{\reftab}[1]{Table~\ref{#1}}
\newcommand{\refsec}[1]{Section~\ref{#1}}
\newcommand{\refappendix}[1]{Appendix~\ref{#1}}

\usepackage{graphicx}
\usepackage{overpic}
\usepackage{wrapfig}

\usepackage[table]{xcolor}
\usepackage{booktabs}

\usepackage{tikz}

\title{\mindmap: Spatial Memory in Deep Feature Maps for 3D Action Policies}

\author{
  \textbf{Remo Steiner}\textsuperscript{*}, \textbf{Alex Millane}\textsuperscript{*}, \textbf{Clemens Volk}\textsuperscript{*}, \textbf{David Tingdahl}\textsuperscript{*},  \textbf{Vikram Ramasamy}\textsuperscript{*}, \textbf{Xinjie Yao}\textsuperscript{*}, \textbf{Peter Du}, \textbf{Soha Pouya}, \textbf{Shiwei Sheng} \\
  \vspace{-1.0em} 
  NVIDIA, Zurich, Switzerland. Santa Clara, California.\\
  \small{\texttt{\{remos,amillane,cvolk,dtingdahl,vramasamy,xyao,peterd,spouya,shiweis\}@nvidia.com}}  \\
}

\begin{acronym}
\acro{TSDF}{Truncated Signed Distance Field}
\acro{VLM}{Vision-Language Model}
\acro{VLA}{Vision-Language-Action}
\acro{VFM}{Vision Foundation Model}
\acro{DDPM}{Denoising Diffusion Probabilistic Model}
\acro{FPN}{Feature Pyramid Network}
\acro{FOV}{Field of View}
\acro{PCA}{Principal Component Analysis}
\end{acronym}

\begin{document}
\maketitle



\begin{abstract}
    End-to-end learning of robot control policies, structured as neural networks, has emerged as a promising approach to robotic manipulation. 
    To complete many common tasks, relevant objects are required to pass in and out of a robot's field of view.   
    In these settings, spatial memory - the ability to remember the spatial composition of the scene - is an important competency.
    However, building such mechanisms into robot learning systems remains an open research problem.
    We introduce \mindmap (Spatial \textbf{M}emory \textbf{in} \textbf{D}eep Feature \textbf{M}aps for 3D \textbf{A}ction \textbf{P}olicies), a 3D diffusion policy that generates robot trajectories based on a semantic 3D reconstruction of the environment.
    We show in simulation experiments that our approach is effective at solving tasks where state-of-the-art approaches without memory mechanisms struggle.
    We release our reconstruction system\footnote{\rurl{github.com/nvidia-isaac/nvblox}}, training code\footnote{\rurl{github.com/nvidia-isaac/nvblox_mindmap}}, and evaluation tasks\footnotemark[2] to spur research in this direction.

\end{abstract}




\begin{figure}[H]
    \centering
    \begin{overpic}[width=0.45\linewidth,t]{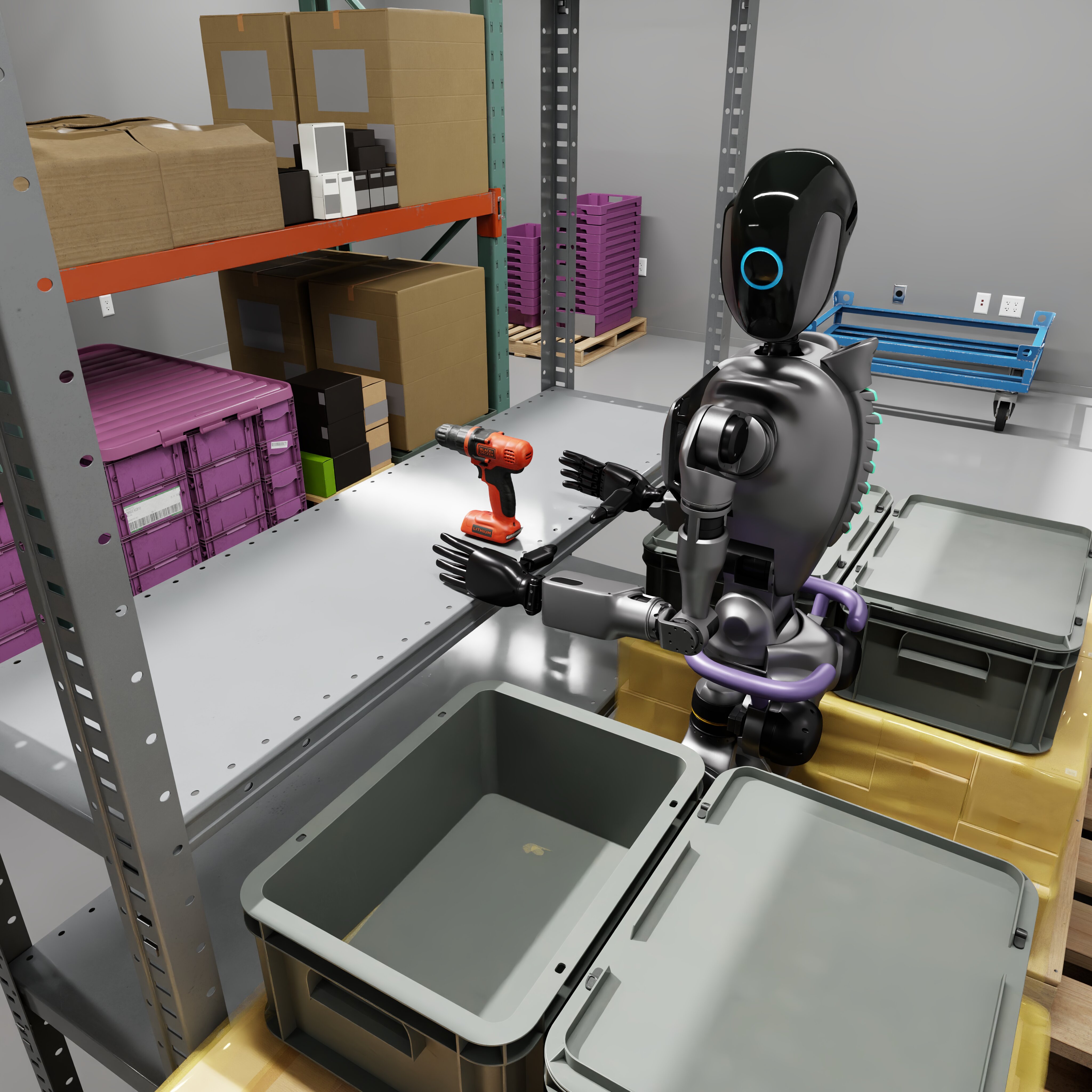}
    \put(100,65){%
        \setlength{\fboxsep}{0pt}
        \setlength{\fboxrule}{2pt}
        \fcolorbox{white}{white}{%
            \includegraphics[width=0.15\linewidth]{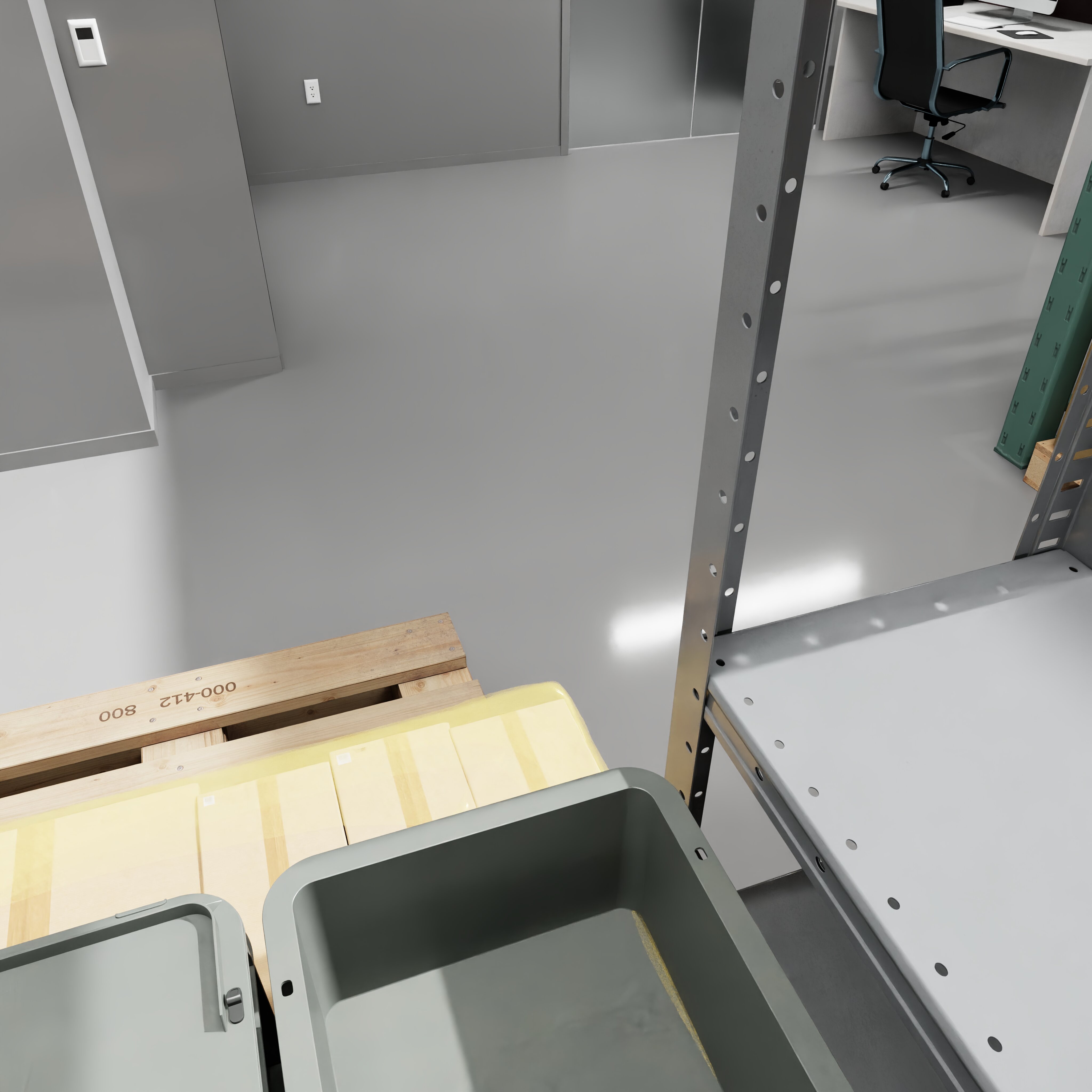}%
        }%
    }
    \end{overpic}
    \includegraphics[width=0.45\linewidth]{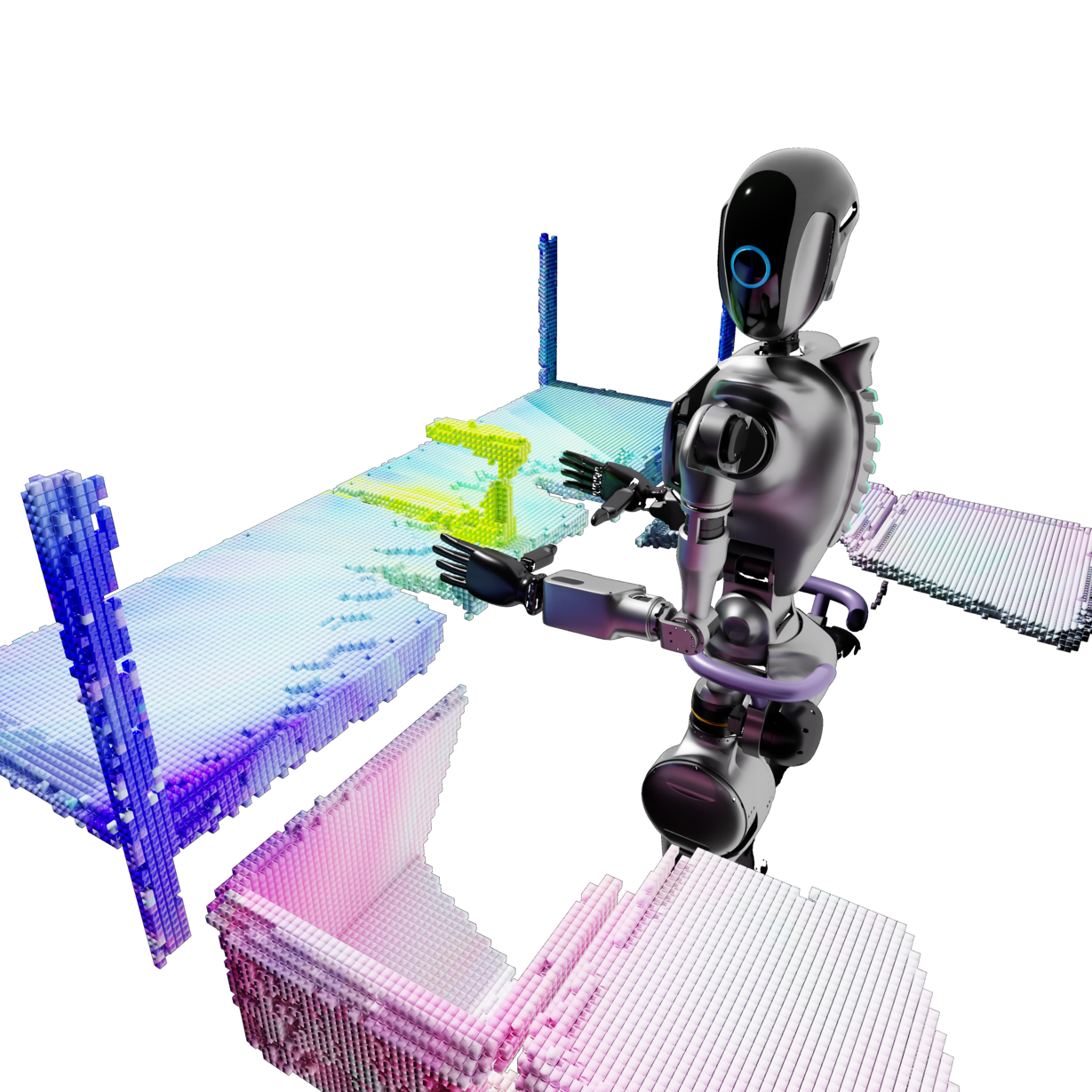}
    \caption{\textbf{Spatial Memory Task: } A humanoid in a simulated industrial space (left) and within a metric-semantic reconstruction built by \mindmap(right) (colored by \acl{PCA}). The robot's first-person view is shown inset. The task requires the robot to transfer the hand drill from the shelf to the open box. The drill and box positions must be discovered by the policy, and both objects cannot be captured in a single view. Therefore, successful task completion requires the policy to remember the spatial layout of the scene. By leveraging the reconstruction, \mindmap generates trajectories that depend on parts of the scene that are outside the robot's current \acl{FOV}.
    }
    \label{fig:teaser}
\end{figure}

\begin{figure}[h]
    \centering
    \includegraphics[width=0.95\linewidth]{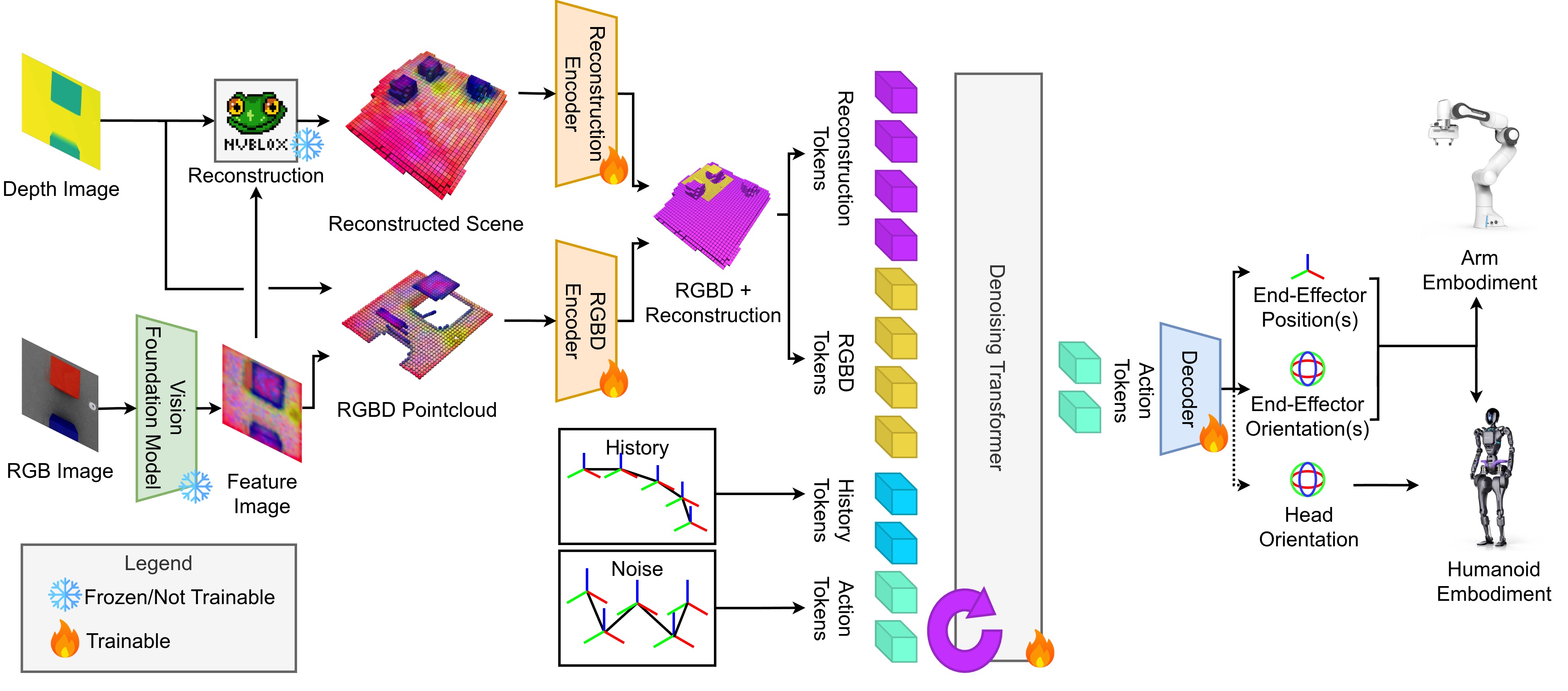}
    \caption{\textbf{Overview of \mindmap}. \mindmap is a \acl{DDPM} that samples robot trajectories conditioned on sensor observations and a reconstruction of the environment. Images are first passed through a \acl{VFM} and then back-projected, using the depth image, to a pointcloud (as in 3D Diffuser Actor~\cite{3d_diffuser_actor}). In parallel, a reconstruction of the scene is built that accumulates metric-semantic information from past observations. The two 3D data sources, the instantaneous visual observation and the reconstruction, are passed to a transformer that iteratively denoises robot trajectories.
    }
    \label{fig:architecture}
\end{figure}

\section{Introduction}
\label{sec:introduction}


Designing generalist robot manipulation policies remains a holy grail of robotics.
Such policies would perform manipulation tasks with a high level of competence and be instructed to do so in natural language.
Recent advances in deep learning, vision, and natural language processing have, for the first time, brought this goal within reach; however, significant challenges remain.

Existing approaches to developing learned manipulation policies generally aim to learn a mapping from sensor observations to robot control signals~\cite{bjorck2025gr00t,black2024pi_0,kim2024openvla,zitkovich2023rt}.
These models typically employ transformer-based architectures to process image and proprioceptive inputs to generate control signals.
Such methods have shown an impressive ability to complete language-guided manipulation tasks.
One limitation of several leading approaches, however, is that the generation of output signals is conditioned on \textit{current} visual observations only.
Such approaches lack spatial memory - the ability to remember the spatial and semantic composition of the scene (see~\cite{cherepanov2025memory} for a taxonomy of robot memory).
This leads to surprising limitations to their capabilities.
Although some methods incorporate temporal information by maintaining a temporal window of past images, these approaches have drawbacks of their own (see \refsec{sec:related_work}).

In this work, we introduce \mindmap, an approach that combines a diffusion policy with a metric-semantic 3D reconstruction of the scene.
\mindmap generates trajectories of 3D end-effector poses in the reconstructed space.
This approach allows the policy to generate actions that depend on parts of the scene that are outside of the camera's current \ac{FOV}.
Our experiments show that, on tasks requiring spatial memory, \mindmap is effective in completing tasks on which several current approaches struggle.



\textbf{Contributions: }
In this paper, we contribute tools for extending 3D manipulation policies with spatial memory.
In particular, we release metric-semantic mapping\footnote{\rurl{nvidia-isaac.github.io/nvblox/pages/torch_examples_deep_features}} in \nvbloxtorch~\cite{millane2024}, our GPU-accelerated reconstruction library\footnotemark[1], in addition to our training code\footnotemark[2], and simulation environments\footnotemark[2] for testing spatial memory.
We demonstrate the efficacy of these tools by extending a state-of-the-art 3D diffusion policy~\cite{3d_diffuser_actor}.
We show that by making changes to the architecture and training, the policy's performance, on challenging tasks that require spatial memory, is significantly improved.



\section{Related Work}
\label{sec:related_work}

Learning robot control policies that map observations directly to robot actions has received considerable recent attention.
Following the success of deep learning in other fields, structuring these policies as neural networks has emerged as a promising approach for building generally intelligent machines.

\textbf{Vision-Language-Action Models:}  Recent robotics research has attempted to replicate the success of large-scale task-agnostic pre-training in other fields, such as language understanding.
RT-1~\cite{brohan2022rt} trained a transformer-based model to produce discrete action tokens on a dataset of 130k demonstrations.
To improve generalization and reasoning abilities, several approaches have sought to incorporate \acp{VLM} into robotic models, the combination termed \ac{VLA} models.
RT-2~\cite{zitkovich2023rt} and OpenVLA~\cite{kim2024openvla} fine-tune \acp{VLM} with robot data, resulting in state-of-the-art zero-shot performance.
These models faced limits in their dexterity due to action-space discretization and execution frequency.
The $\pi_0$~\cite{black2024pi_0} model addressed these limitations, using a diffusion-based action head~\cite{chi2023diffusion} to represent continuous distributions over action-space, and to produce high-frequency output.
GR00T N1~\cite{bjorck2025gr00t} suggests a flow-matching-based \ac{VLA} trained on varied data sources.
Many recent works have sought to improve \ac{VLA} models through improved action tokenization~\cite{pertsch2025fast}, action-chunking~\cite{zhao2023learning,black2025real}, and multi-step instruction following~\cite{shi2025hi}, among others.

\textbf{3D Manipulation Models:} In parallel, efforts have been made to train models that utilize 3D sensor data.
Perceiver-Actor~\cite{shridhar2023perceiver} voxelizes an RGB-D pointcloud and uses a transformer to produce language-conditioned goals.
RVT~\cite{goyal2023rvt} represents the 3D scene through several virtual views, leading to dramatically improved training times. 
3D Diffuser Actor~\cite{3d_diffuser_actor} represents the scene as a set of featurized 3D points, and processes them using 3D relative attention to produce continuous actions.
At the time of writing, policies consuming 3D data have not typically undergone large-scale pre-training.
FP3~\cite{yang2025fp3} represents an early attempt to scale up a 3D policy, using the DRIOD~\cite{khazatsky2024droid} dataset.

\textbf{Reconstruction for Manipulation: } Several works have investigated the use of reconstructions in manipulation policies.
LERF-TOGO~\cite{rashid2023language} and SplatMover~\cite{shorinwa2024splat} build metric-semantic maps upon which grasp points are predicted, using NERFs and Gaussian splats respectively.
In contrast, \mindmap follows an end-to-end approach, diffusing robot trajectories directly from a reconstruction, without intermediate prediction of grasps.
GNFactor~\cite{ze2023gnfactor} uses several external cameras to build a 3D voxel grid of \ac{VFM} features, which are then processed by a transformer to produce voxelized actions.
The reconstruction, however, is built from views of the scene at a single timestep.
In contrast, our results are generated using a single ego-centric camera that accumulates prior views of the scene to provide past information to the network.

\textbf{Memory:} One limitation of many \acp{VLA} and 3D models is that they produce actions based on the \textit{current} observation.
As we shall show, this is a significant limitation, even on seemingly trivial tasks.
A recent work SAM2ACT~\cite{fang2025sam2act}, addresses the issue of spatial memory in manipulation policies.
The authors propose adding a memory bank to RVT2~\cite{goyal2024rvt}, feeding back prior observations into the policy.
The authors demonstrate state-of-the-art performance on tasks requiring spatial memory.
However, as the authors note, the approach has several shortcomings.
SAM2ACT has a fixed-length memory that requires per-task tuning.
The model's recurrent nature requires a specialized training procedure.
In contrast, the approach proposed in \mindmap has no explicit temporal limits. Past information is aggregated spatially, rather than stored in a temporal buffer, and so the computational requirements remain bounded given a bounded volume of space.
Furthermore, the approach is not recurrent, and so can be plugged directly into a standard diffusion policy training pipeline.


\section{Problem Statement}
\label{sec:problem_statement}

Given a sequence of observations $\mathcal{O} = \{ \mathbf{o}_i \}_{i=0}^{t} $ we aim to find a policy $\pi$ that outputs a robot action $\mathbf{a}_t$ such that $\mathbf{a}_t = \pi(\mathbf{o}_0, \mathbf{o}_1, ..., \mathbf{o}_t)$.
Our observations $\mathbf{o}_i$ take the form of $\mathbf{o}_i = \{\mathcal{I}_i^j, \mathcal{D}_i^j, \mathcal{S}_i\}_{j=0}^N$, for $N$ cameras, where $\mathcal{I}_i^j$ are RGB images, $\mathcal{D}_i^j$ are corresponding posed depth images, and $\mathcal{S}_i$ is the robot state $\mathcal{S}_i = \{\mathbf{p}_i^k, \mathbf{q}_i^k, c_i^k, \gamma_i\}_{k=0}^M$, for $M$ end-effectors.
We consider several robot embodiments, but in general, the robot state $\mathcal{S}_i$ is a composition of the 3D positions $\mathbf{p}_i^k \in \mathbb{R}^3$, rotations $\mathbf{q}_i^k \in \text{SO(3)}$, the closedness $c_i^k \in \{0,1\}$ of one or more robot end-effectors, and for humanoid embodiments, the head yaw $\gamma_i \in (-\pi, \pi]$.
Our action $\mathbf{a}_i$ lives in the same space as our state $\mathcal{S}_i$, i.e. we command end-effector poses, closedness, and head yaw.
Our policy $\pi$ is a deep neural network which we learn from human demonstrations consisting of observation-action pairs $\mathcal{T} = \{(\mathbf{a}_0, \mathbf{o}_0), (\mathbf{a}_1, \mathbf{o}_1), ..., (\mathbf{a}_T, \mathbf{o}_T)\}$.
We build a reconstruction $\mathcal{R}_t$ by accumulating past visual observations $\mathcal{R}_t(\mathcal{I}_0,...,\mathcal{I}_t, \mathcal{D}_0,...,\mathcal{D}_t)$.
Our policy depends on the current observations directly, a finite sequence of $K$ past states, and
on past visual observations through the reconstruction $\mathbf{a}_t =  \pi(\mathcal{I}_t^j, \mathcal{D}_t^j,\mathcal{S}_{t-K},...,\mathcal{S}_t, \mathcal{R}_t)$



\section{Method}
\label{sec:method}

In this section we describe our approach, firstly describing our extensions to 3D Diffuser Actor~\cite{3d_diffuser_actor} (\refsec{sec:method_network}), and then explaining how we build reconstructions (\refsec{sec:reconstruction}). See \reffig{fig:architecture} for an overview.

\subsection{Network Architecture}
\label{sec:method_network}

Our approach follows recent work~\cite{3d_diffuser_actor, chi2023diffusion, black2024pi_0} and structures our policy as a denoising transformer that generates robot actions based on observations of the scene.
In particular, we extend 3D Diffuser Actor~\cite{3d_diffuser_actor}, which iteratively denoises an end-effector trajectory, conditioned on posed RGB-D images. In the following, we highlight the key differences between \mindmap and 3D Diffuser Actor.

\textbf{Reconstruction tokens:} 
\Mindmap's diffusion transformer takes as input RGB-D images and a featurized reconstruction, in the form of 3D vertices extracted from a reconstructed mesh (see Section~\ref{sec:reconstruction}).
This allows the network to attend to both the current RGB-D observation and the reconstruction, which aggregates past observations.
We found that this approach led to better results than providing the reconstruction alone (see \refsec{sec:result}).
The reconstruction is continuously updated as new images arrive.

The featurized RGB-D image and the reconstruction are passed through separate encoders to project them from \ac{VFM} feature dimension to the token embedding dimension (see \reffig{fig:architecture}).
Reconstruction and RGB-D tokens are then concatenated and passed through cross and self-attention layers, as in 3D Diffuser Actor.
We found that the use of separate encoders led to higher performance than passing both sets of points through a joint encoder.
This makes intuitive sense: it allows attention mechanisms to differentiate tokens originating from instantaneous observations and those coming from the reconstruction.

\textbf{\ac{VFM} Features:} Diffuser Actor uses a pre-trained CLIP ResNet50 image encoder~\cite{clip} combined with a trainable \ac{FPN}~\cite{lin2017feature} for feature extraction.
The reconstruction process in \mindmap is non-differentiable and as a result gradients are unable to flow back to the image encoder. 
We therefore replace CLIP+FPN with a frozen pre-trained \ac{VFM}, AM-RADIO~\cite{ranzinger2024radio}.

\textbf{Bimanual embodiments:}
We extend 3D Diffuser Actor, which was designed to control a single robotic arm, for bimanual manipulation tasks using a humanoid robot.
We therefore modify the model from predicting single end-effector poses and closedness to (optionally) predict bimanual end-effector poses and closedness.
We concatenate the past states of multiple end-effectors to form the proprioceptive history, and we modify the prediction heads in the network to predict the next states for multiple end-effectors (as suggested in~\cite{ke2024bi3d}).

\textbf{Controlling head orientation:}
We additionally allow the policy to control the head orientation of humanoid robots.
This allows \mindmap to complete tasks in which not all task-relevant objects can be held in a single view of the scene.
In such situations, the robot must gather information from several views in order to complete the task.
To achieve this, we add an additional decoder for the head orientation.
In training, the head orientation is supervised by the tele-operator's head orientation, captured by a virtual reality device (see \refsec{sec:implementation}).

\subsection{Reconstruction}
\label{sec:reconstruction}

We compute a reconstruction of the scene from all past robot observations using the publicly available \nvbloxtorch library~\cite{millane2024}, which we extend for metric-semantic mapping in PyTorch. This library fuses posed RGB-D sensor data into a \ac{TSDF} in real-time. For each incoming \mbox{RGB-D} frame, \nvbloxtorch projects the 3D grid into the depth image and updates the distance values and weights of affected voxels (described in~\cite{izadi2011kinectfusion}).
Figure~\ref{fig:reconstructions} (Appendix~\ref{appendix:reconstructions}) shows reconstructions for tasks introduced in \refsec{sec:result}.

\textbf{Geometry:} From the distance field, we extract a representation of the 3D surface.
In particular, \nvbloxtorch applies the marching cubes algorithm \cite{lorensenC87} to compute a mesh that represents the zero-level isosurface of the distance field. In this work, we only keep the mesh vertices, i.e. triangle and normal data are discarded. The result is a dense point cloud, build from the fusion of previous visual observations.

\textbf{Features:} 
To generate a metric-semantic representation of the environment, we also fuse \ac{VFM} image features into the reconstructed voxel map. In particular, we extract 2D feature maps $\mathcal{F}_i$ from the incoming RGB images $\mathcal{I}_i$, using a pre-trained \ac{VFM} $\phi$:
\begin{equation}
\mathcal{F}_i = \phi(I_i), \quad \mathcal{F}_i \in \mathbb{R}^{h \times w \times f}
\end{equation}
where $f$ is the channel depth of the feature produced by the \ac{VFM}. The feature associated with each voxel is updated by projecting the voxel center $\mathbf{p} \in \mathbb{R}^3$ into the feature map and reading the feature vector at the projected image point:
\begin{equation}
\mathbf{f}_i = \mathcal{F}_i[\Pi({\mathbf{p}})], \quad \mathbf{f}_i \in \mathbb{R}^f
\end{equation}
Here, $\Pi : \mathbb{R}^3 \to \mathbb{R}^2$ is the camera projection function, and $[.]$ denotes nearest-neighbour pixel lookup.
We found that simply overwriting the existing voxel feature during updates yields similar results as to fusing the incoming feature with the existing one (see \refsec{sec:ablations}). Similar to the TSDF reconstruction, we handle occlusions by only updating voxels in a narrow truncation band around non-occluded surfaces (set to $\pm 4$\,voxels in our experiments). Finally, the mesh vertices are featurized by looking up their closest feature vector in the voxel map.
Appendix~\ref{appendix:nvblox_torch} gives implementation details about achieving this with \nvbloxtorch.







\section{Implementation}
\label{sec:implementation}

In this section, we provide details about the implementation of our method.


\begin{figure}[h]
    \centering
    \includegraphics[width=0.22\linewidth]{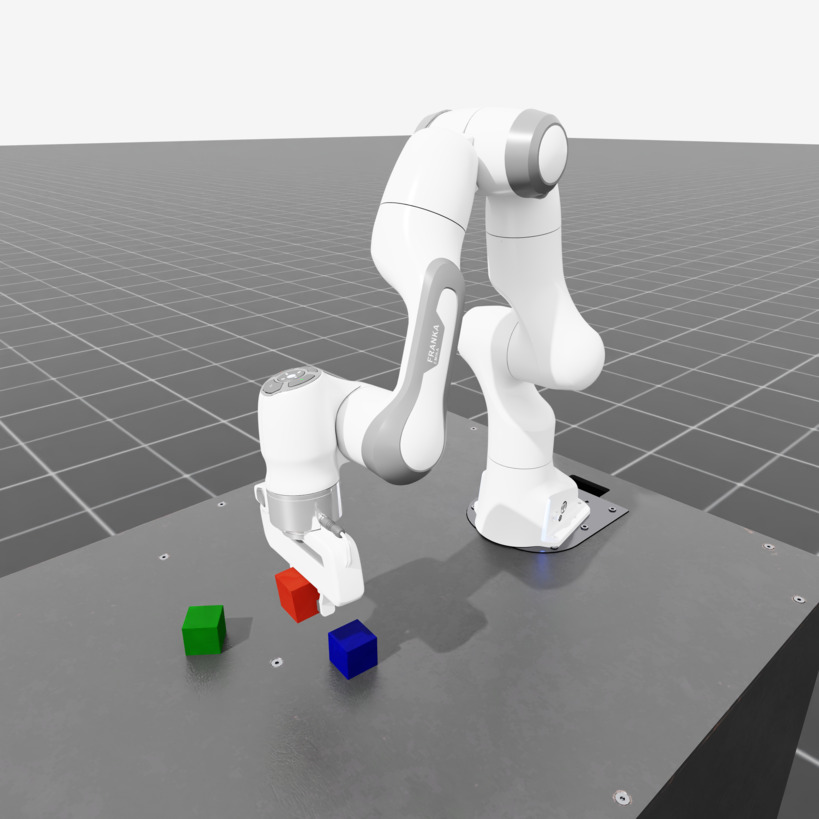}
    \includegraphics[width=0.22\linewidth]{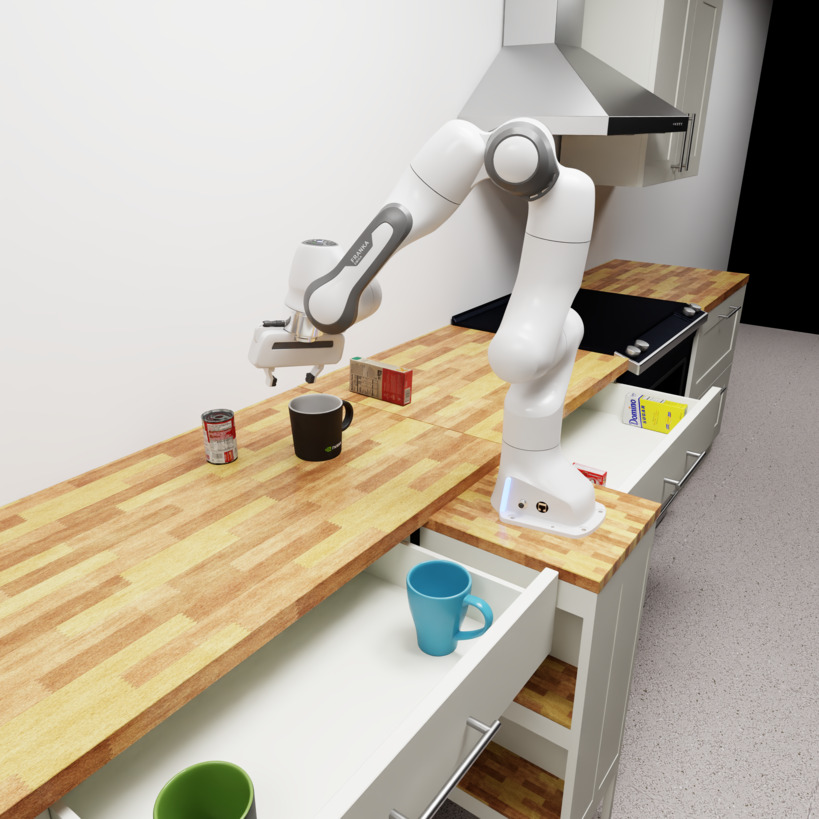}
    \includegraphics[width=0.22\linewidth]{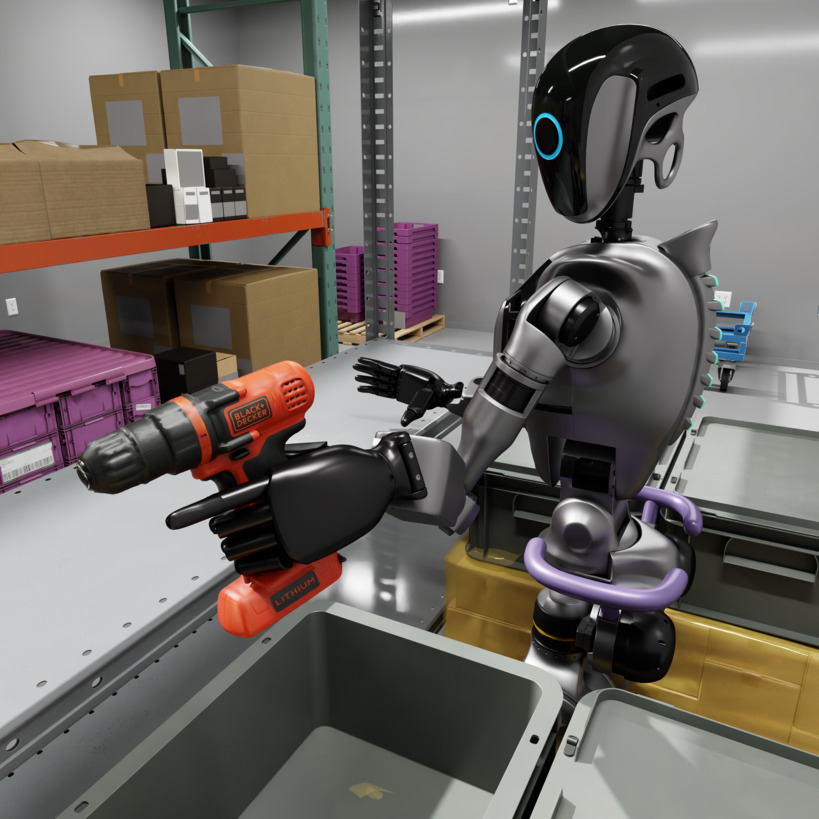}
    \includegraphics[width=0.22\linewidth]{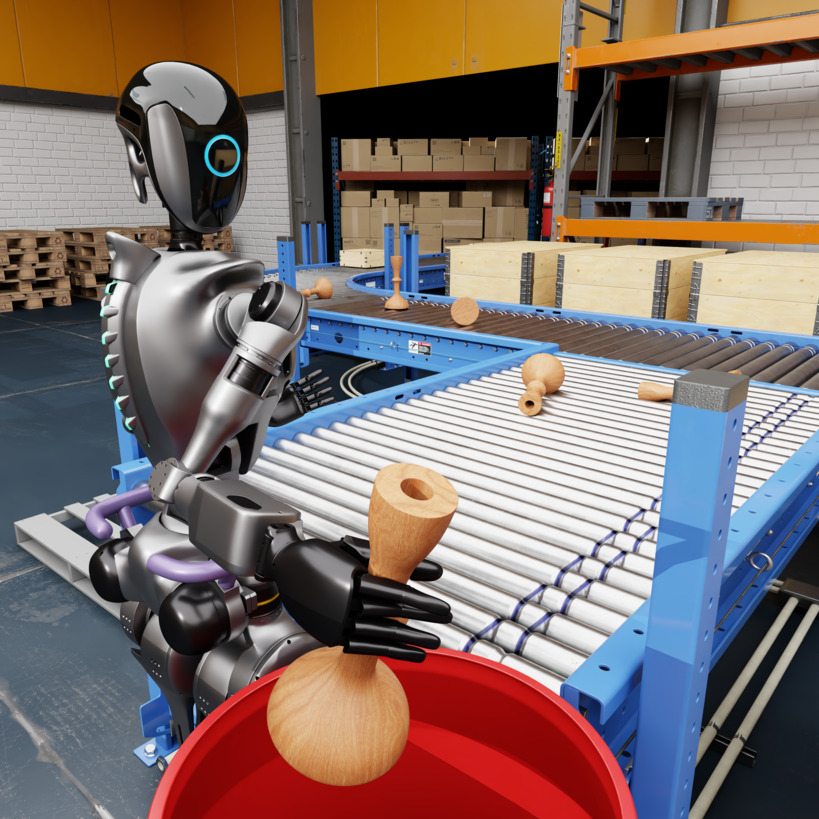}
    \caption{Environments introduced to evaluate policies' spatial memory. From left to right: \textbf{\cubestacking:} stack three cubes (initial cube positions are randomized), \textbf{\mugindrawer} move mug into drawer containing mugs (positions of objects on kitchen counter are randomized and the destination drawer position is permuted), \textbf{\drillinbox:} put hand drill into open box (drill position is randomized and open/closed boxes are permuted), \textbf{\stickinbin:} put candlestick into bin (stick and bin positions are randomized). In all tasks, policies are provided a single ego-centric camera view from which the entire task space cannot fit into the \ac{FOV}.
}
    \label{fig:evaluation_environments}
\end{figure}

\textbf{Demonstration Data Collection: } We simulate several tasks in IsaacLab~\cite{mittal2023orbit} to evaluate \mindmap (see \refsec{sec:result}).
We collect demonstration trajectories through teleoperation using IsaacLab Mimic\footnote{\rurl{isaac-sim.github.io/IsaacLab/v2.1.0/source/overview/teleop_imitation.html}} (based on MimicGen~\cite{mandlekar2023mimicgen}), an Apple Vision Pro for the humanoid robot, and a space-mouse for the robot arm.
The human demonstration trajectories are multiplied to generate a larger dataset.
For each task, we train on 100 trajectories and evaluate on 100 distinct randomizations.


\begin{figure}
    \centering
    \includegraphics[width=0.40\linewidth]{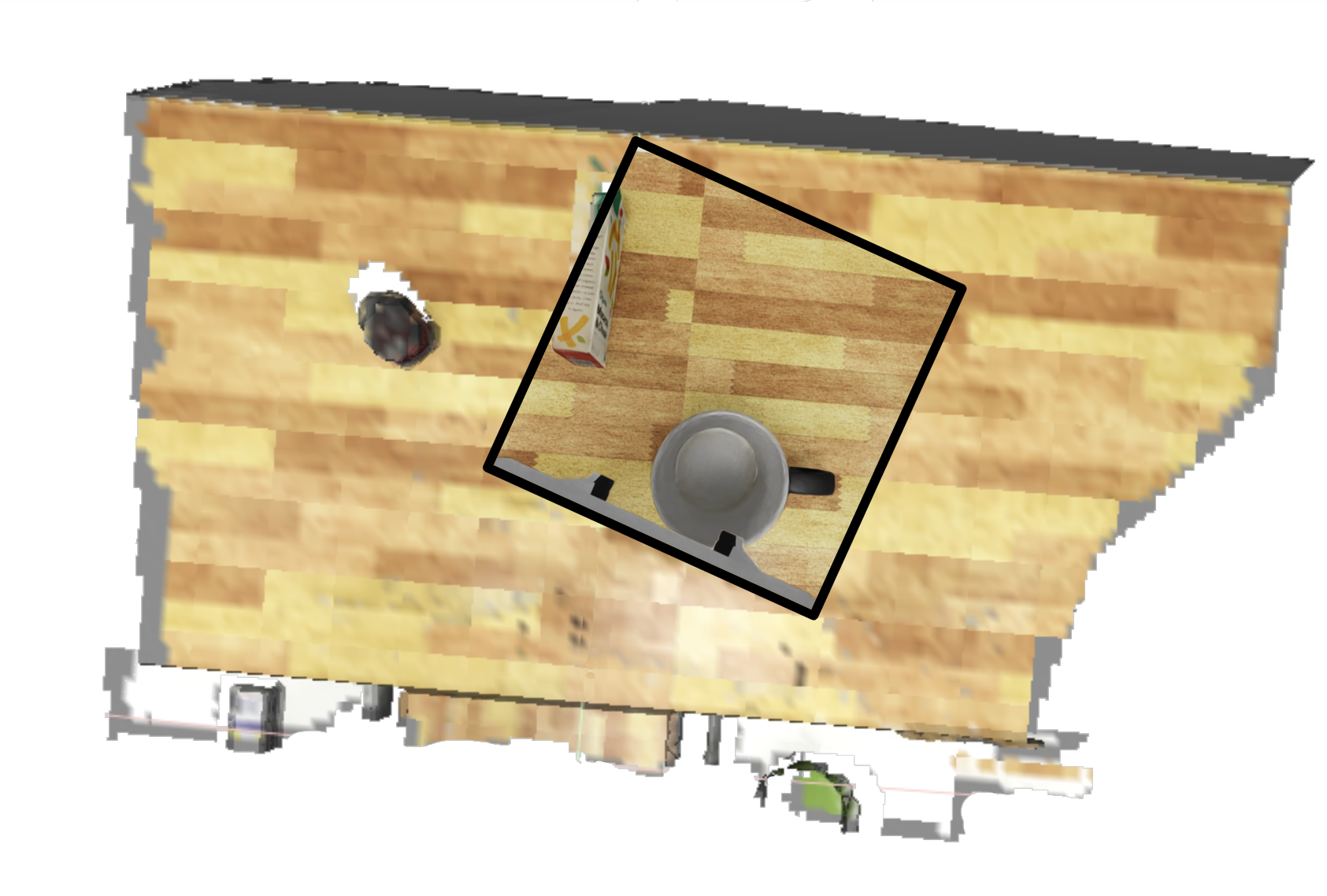}
    \begin{tikzpicture}
    \node[anchor=south west,inner sep=0] (img) at (0,0)
      {\includegraphics[width=0.40\linewidth]{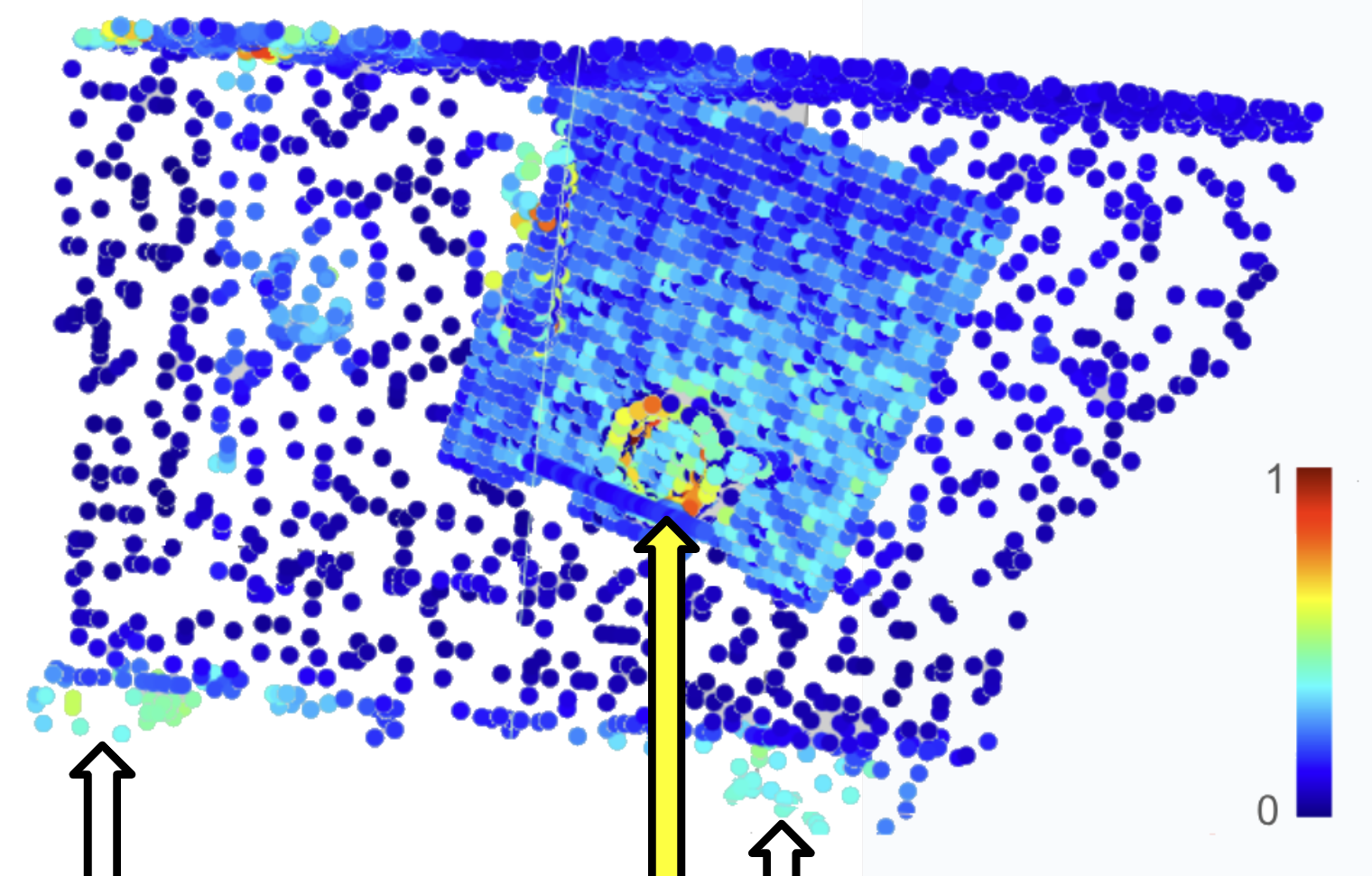}};
    \end{tikzpicture}    
    \caption{\textbf{Attention Visualization: } Top-down visualization of 3D attention weights (right) and reconstruction (left) for the \mugindrawer task. The inset shows the current camera view.  Extrema appear in regions of interest to the task, such as the mug (yellow arrow) and the drawers in the bottom left/right (white arrows). The high concentration of points in the center is generated by the current view of the camera, while points outside this region are from the reconstruction.
    }
    \label{fig:attention_weights}
\end{figure}

\textbf{Reconstruction Data Generation: }
During training, we select a random timestamp in the demonstration trajectory and attempt to predict the next keypose based on the RGBD observation, the state history, and the reconstruction.
We therefore need random access to reconstructions associated with each timestamp in the demonstration trajectories.
To achieve this, we perform mapping for each demonstration trajectory and save a per-timestamp reconstruction before training.
Producing a dataset of 100 trajectories (from 10 human demonstrations), including running the IsaacLab, RTX raytracing, and \nvbloxtorch reconstruction, takes 4 hours on a single L40 GPU node, producing approximately 3000 reconstructions.
\ac{TSDF} reconstruction is performed at 1\,cm voxel resolution.

\textbf{Training: } 
Training runs are performed on a 2-GPU H100 node for 150k iterations, taking approximately 2 days.



\section{Results}
\label{sec:result}

In this section, we aim to validate the hypothesis of this paper, that \mindmap improves performance on tasks that require spatial memory.

\textbf{Evaluation Environments: } Existing benchmarks like RLBench~\cite{james2019rlbench} focus on table-top manipulation tasks in which all task-relevant objects remain in view at all times.
These tasks do not require spatial memory for completion because the entire state of the task can be determined from a single view.

We therefore introduce four challenging tasks on which to evaluate policies for spatial memory use (see \reffig{fig:evaluation_environments}).
We restrict policies to ego-centric observations of the scene: the wrist camera for robot arm tasks, and to a head camera for humanoid tasks.
An ego-centric camera is practical, as the robot is freed from a reliance on external infrastructure, which will become increasingly important as robots are expected to mix manipulation with movement through the environment.
In our tasks, the robot is unable to see all task-relevant objects within its field of view at all times (see \reffig{fig:environment_views}).
As a consequence, the policy needs to remember the spatial layout of the scene to complete the task with a high success rate (see \refappendix{appendix:task_descriptions} for descriptions of the tasks).
While this type of task is somewhat novel for manipulation policy evaluation, it is very common in everyday life; humans are frequently required to reason about out-of-view objects.

\textbf{Baselines: } We compare \mindmap with 3D Diffuser Actor~\cite{3d_diffuser_actor}.
For humanoid tasks, we also compare against GR00T N1~\cite{bjorck2025gr00t}. To match \mindmap, we modify 3D Diffuser Actor to utilize AM-RADIO~\cite{ranzinger2024radio} features rather than CLIP~\cite{clip}, which we found to increase performance. 
We also compare to a version of 3D Diffuser Actor that is provided with an external camera to remove the requirement for memory on our tasks.
\mindmap and 3D Diffuser Actor are trained from scratch, while GR00T N1 is fine-tuned on each task.
We attempted to fine-tune GR00T N1 on the robot arm tasks, but were unable to achieve non-zero success rates, likely because ego-centric-only robot arm tasks are not in its pretraining data. We omit these results.

\begin{table}
    \centering
    \tiny
    \begin{tabular}{c>{\columncolor{green!20}}ccccc}
    \toprule
    {} & {\cellcolor{white}} & \multicolumn{4}{c}{\textbf{Task}}  \\
    \cmidrule{3-6}
       \textbf{Method} & \textbf{Average} & \textbf{\cubestacking} & \textbf{\mugindrawer} & \textbf{\drillinbox} & \textbf{\stickinbin} \\
    \midrule
         Mindmap                                   &  \textbf{76\%} (\textbf{80\%}) & \textbf{47\%} & \textbf{97\%} & \textbf{78\%} & \textbf{82\%}   \\
         3D Diffuser Actor \cite{3d_diffuser_actor}  &  20\% (18\%) & 0\%           & 46\%          & 21\%          & 14\%   \\
         GR00T N1~\cite{bjorck2025gr00t}           &  - \quad (54\%)   & -             &  -            & 46\%          & 62\%   \\
    \midrule
         Privileged (external cam) 3D Diffuser Actor \cite{3d_diffuser_actor}  &  85\% (85\%) & 74\%  & 97\%  & 86\%  & 83\%   \\
    \bottomrule
    \end{tabular}
    \caption{\textbf{Key Findings - Evaluation in Simulation}. \Mindmap is compared against 3D Diffuser Actor~\cite{3d_diffuser_actor} and GR00T N1~\cite{bjorck2025gr00t} in simulated tasks that require spatial memory to complete with a high success rate. We also evaluate a method that uses an external camera as privileged information. The bracketed average is over humanoid tasks only.}
    \label{tab:evaluation_isaac_lab}
\end{table}


\subsection{Key Findings}

\reftab{tab:evaluation_isaac_lab} shows quantitative results comparing \mindmap to the baseline methods.
\mindmap achieves an average success rate of 76\%, an improvement of 56\% (absolute) over 3D Diffuser Actor and 26\% over GR00T N1 (on humanoid tasks).
Further, \mindmap performs only slightly (9\% absolute) worse than the method that is provided with privileged information.
These results, taken together, indicate the efficacy of \mindmap at solving tasks that require spatial memory.

Three of the four tasks (\mugindrawer, \drillinbox, and \stickinbin) involve a binary decision about out-of-view objects.
A policy without spatial memory is reduced to guessing between the two options seen in the training data.
The results, therefore, align with expectations: allowing for the random decision, GR00T N1 achieves close to the best possible performance.
Qualitatively, observation of policy roll-outs confirms this: the policy is very effective at picking up objects; however, it often ($\sim$50\% of cases) makes the wrong binary decision.
By contrast, \mindmap rarely makes the wrong decision, and failures typically originate from object pick-up.

Figure~\ref{fig:attention_weights} shows the attention weights for the \mugindrawer task from the first cross-attention layer in \mindmap.
The figure indicates that network assigns a high weight to the mug to be transported, and both of the drawers, one of which is the target location.
This aligns with intuition: the network attends to task-relevant parts of the scene.
Note that only the mug is within the current camera view.
The assignment of high weight to points outside of the current camera view also indicates the importance of the reconstruction in completing the task.

Lastly, \groot N1 is outperformed by \mindmap by 26\% (absolute).
It is pre-trained on a large dataset and is a much larger model than \mindmap ($\sim$1B trainable parameters, plus $\sim$1B in the frozen VLM vs. \mindmap's $\sim$3M trainable, plus $\sim$100M frozen in the image encoder).
We believe that these results indicate the potential for improving \acp{VLA} through spatial memory mechanisms.

\subsection{Ablations and variations}
\label{sec:ablations}

Table~\ref{tab:ablation} shows the results of varying various design decisions in our method, evaluated on our robot arm tasks.

\textbf{Reconstruction only:} We restrict our method to access the reconstruction only by removing the RBGD pointcloud input to our model.
This leads to a 9\% lower success rate.
Qualitatively, we observe an increased frequency of failure during pick-up. This aligns with intuition: the wrist camera provides high-resolution information during object pick-up, which is likely important for accurate grasping.

\textbf{No \ac{VFM}:} RADIO-AM features are replaced with RGB triplets extracted from the images.
This leads to a 27\% lower success rate.
The relative reduction in success is less pronounced for \cubestacking than for \mugindrawer, likely due to the distinct RGB colors of the cubes providing sufficient information for the model in most cases. In general, compared to semantically rich features like RADIO-AM, raw RGB does not take any contextual or semantic information into account, and its values strongly depend on lighting conditions and viewing direction.

\textbf{Feature blending:} During reconstruction, our baseline method overwrites existing feature vectors with the most recently extracted ones. As an alternative, we explored fusing new measurements with old ones. Here, we update the feature associated with each voxel by applying an exponential filter:
\begin{equation}
\mathbf{f}_{\text{voxel}}(\mathbf{p}) \leftarrow \alpha \cdot \mathcal{F}[\Pi(\mathbf{p})] + (1 - \alpha) \cdot \mathbf{f}_{\text{voxel}}(\mathbf{p})
\end{equation}
We use $\alpha = 0.1$, i.e., a new measurement contributes $10\%$ to the updated value. We found that this modification leads to no significant change in performance. 

\begin{table}
    \centering
    \tiny
    \begin{tabular}{c>{\columncolor{green!20}}ccc}
    \toprule
    {} & {\cellcolor{white}} & \multicolumn{2}{c}{\textbf{Task}}  \\
    \cmidrule{3-4}
       \textbf{Ablation} & \textbf{Average} & \textbf{\cubestacking} & \textbf{\mugindrawer} \\
    \midrule
        \mindmap (baseline) & 72\% & 47\% & 97\% \\
        Reconstruction only            & 63\% & 33\% & 93\%\\
        No \ac{VFM}         & 45\% & 31\% & 59\%\\
        Feature blending    & 74\% & 50\%  & 98\% \\
    \bottomrule
    \end{tabular}
    \caption{\textbf{Ablations and Variations}. Variations of design parameters of \mindmap and their corresponding success rates on the robot arm tasks introduced in \refsec{sec:result}. \textit{Reconstruction only}: removal of RGBD observations. \textit{No \ac{VFM}}: features replaced with RGB triplets. \textit{Feature blending}: blends \ac{VFM} features over time, rather than taking the latest observed feature.}
    \label{tab:ablation}
\end{table}

\subsection{Limitations}
\label{sec:limitations}
Our method has several limitations.
Firstly, our model is small (3\,million trainable parameters), is trained on a small dataset, and in a task-specific regime.
Policies of this kind~\cite{3d_diffuser_actor, chi2023diffusion, ze2023gnfactor, goyal2023rvt, shridhar2023perceiver, fang2025sam2act, goyal2024rvt} are convenient to perform research on, however, do not in general, generalize out of their training environment.
It is an interesting research direction to scale up \mindmap to a larger dataset such as DROID~\cite{khazatsky2024droid}. 
Secondly, our model produces end-effector keyposes as output. 
Keypose extraction from VR teleop data is non-trivial and task-specific.
Altering the model to predict trajectories using action-chunking~\cite{zhao2023learning}, as is common in \acp{VLA}, has the potential to remove the limiting step.
Lastly, our reconstruction process is non-differentiable. The result is that we store a full \ac{VFM} feature \textit{per-voxel}, which requires substantial amounts of storage during training and memory during inference. There is an opportunity, with a differentiable reconstruction process, to do learned dimensionality reduction before reconstruction to reduce memory consumption.


\section{Conclusions}
\label{sec:conclusion}

In this paper, we present \mindmap, a manipulation policy that diffuses robot trajectories from a reconstruction of the observed scene.
We showed that tasks involving spatial memory are challenging for methods that compute trajectories based on the current observation only.
\Mindmap is able to utilize past information, in the form of the metric-semantic reconstruction, in order to complete tasks that involve reasoning about out-of-view objects.
The result is that \mindmap significantly improves performance on spatial memory evaluations.
We contribute our tools for metric-semantic mapping and for training reconstruction-based diffusion policies to spur further research in this direction.
We foresee a growing importance of spatial memory as learned manipulation policies move beyond the tabletop tasks, in particular to tasks that combine locomotion and manipulation.


\clearpage


\acknowledgments{We would like to thank the 3D Diffuser Actor~\cite{3d_diffuser_actor} authors for open-sourcing their code, in particular Nikolaos Gkanatsios for his generosity with his time, and for the fruitful discussions about 3D manipulation policies.}


\bibliography{references}  

\newpage

\section{Appendix}
\label{sec:appendix}


\subsection{Example Reconstructions}
\label{appendix:reconstructions}

\begin{figure}
    \centering
    \includegraphics[width=0.40\linewidth]    {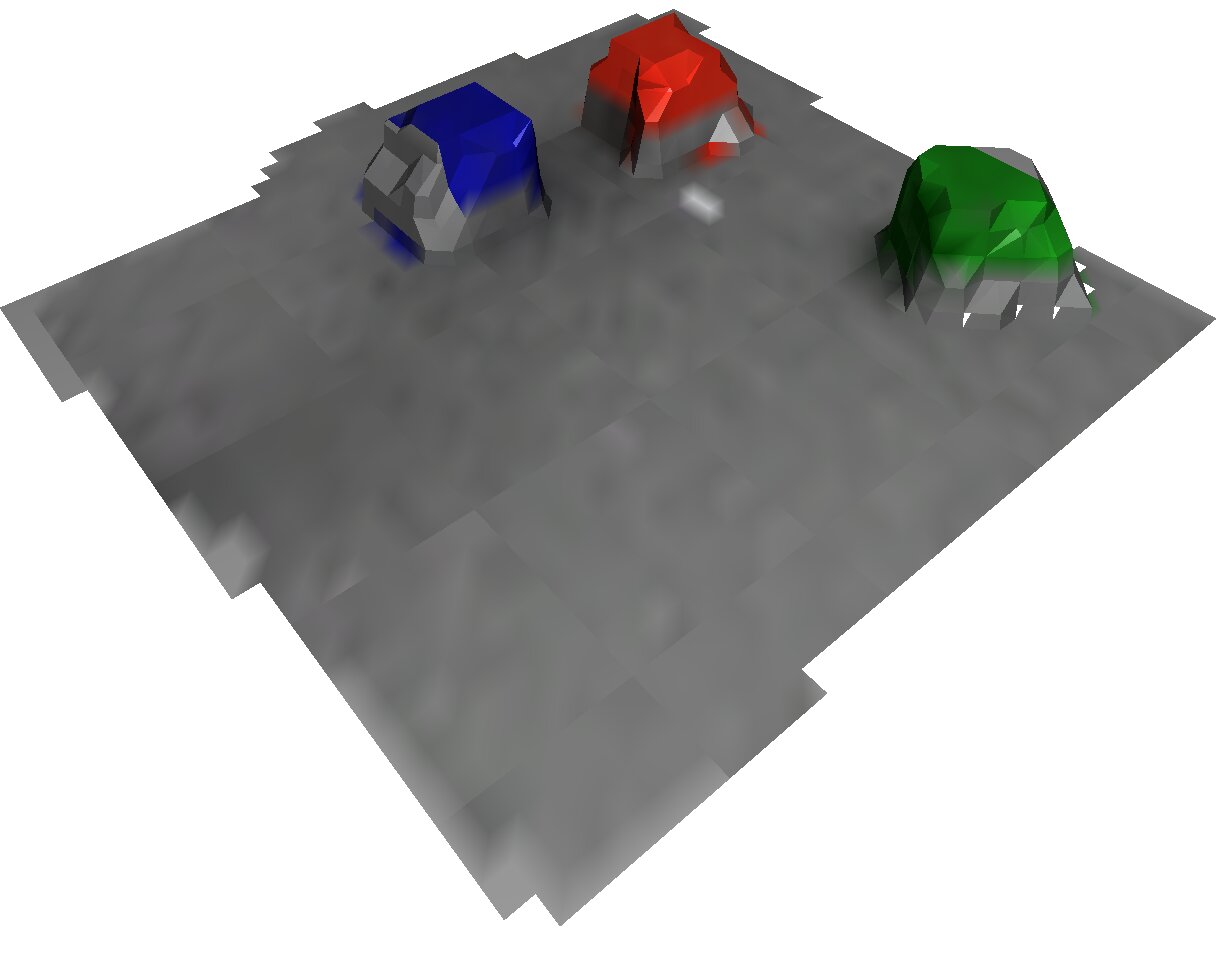}
    \hspace{2.0em}
    \includegraphics[width=0.40\linewidth]{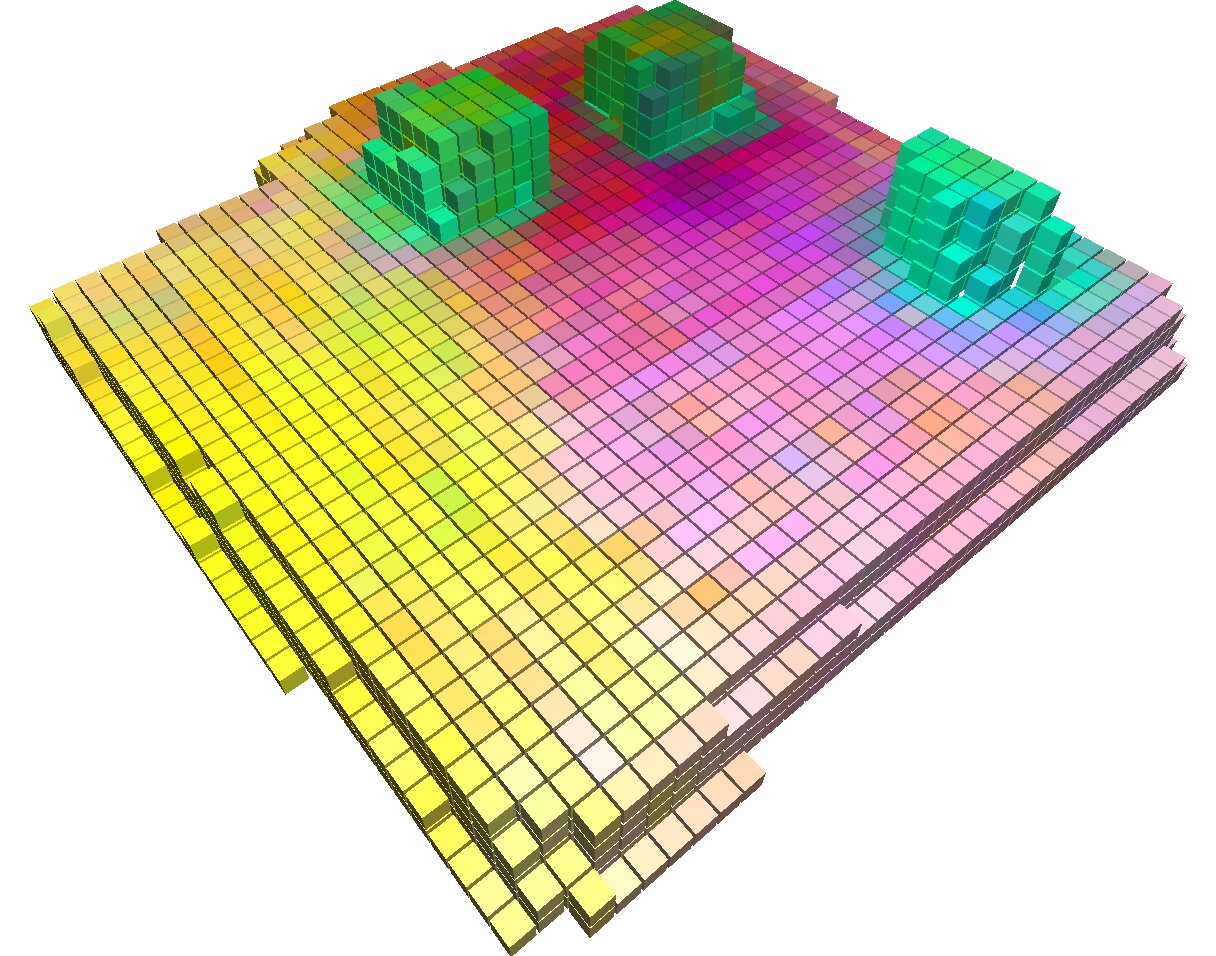}
    \\[5ex]  
    \includegraphics[width=0.40\linewidth]{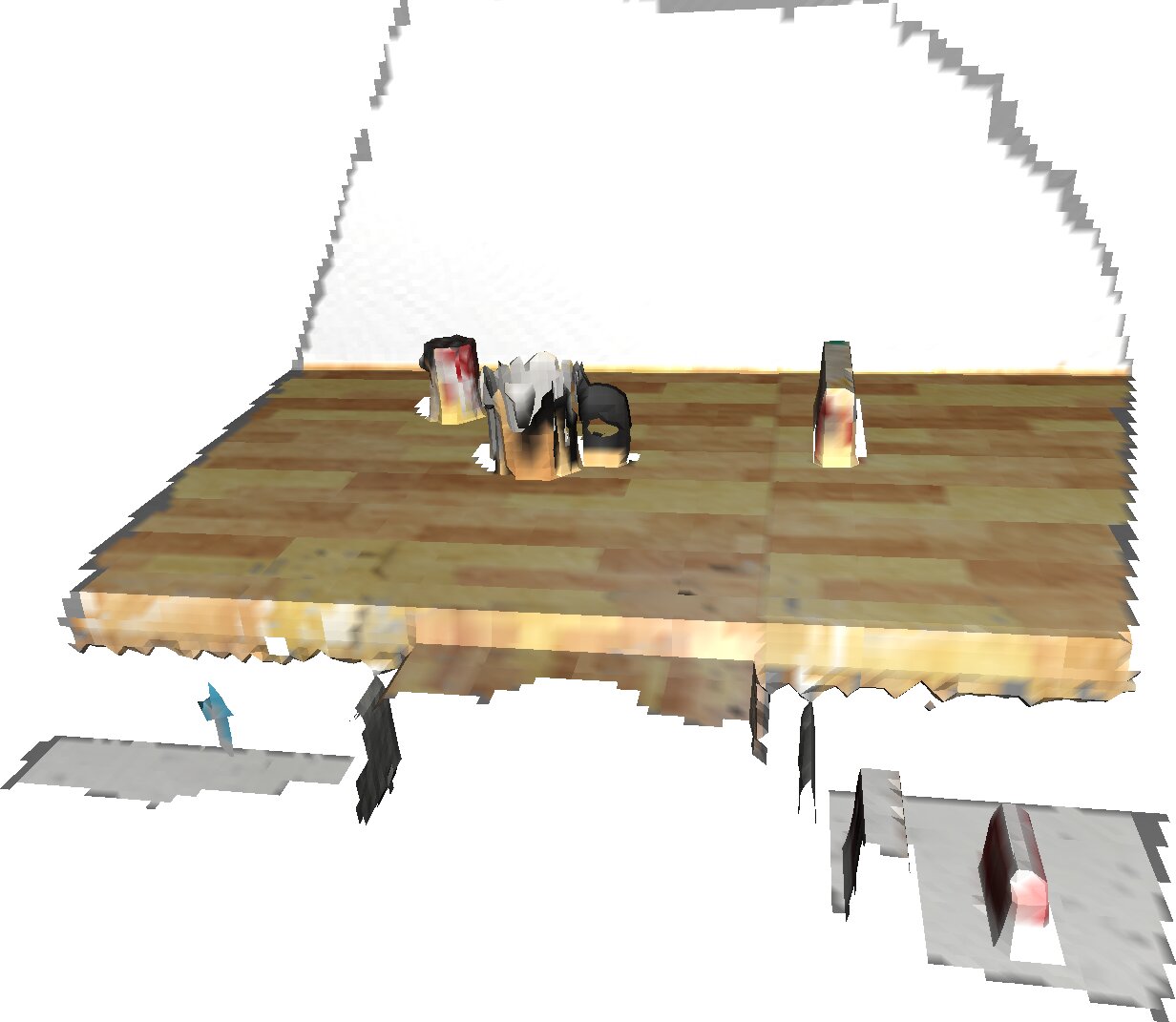}
    \hspace{2.0em}
    \includegraphics[width=0.40\linewidth]{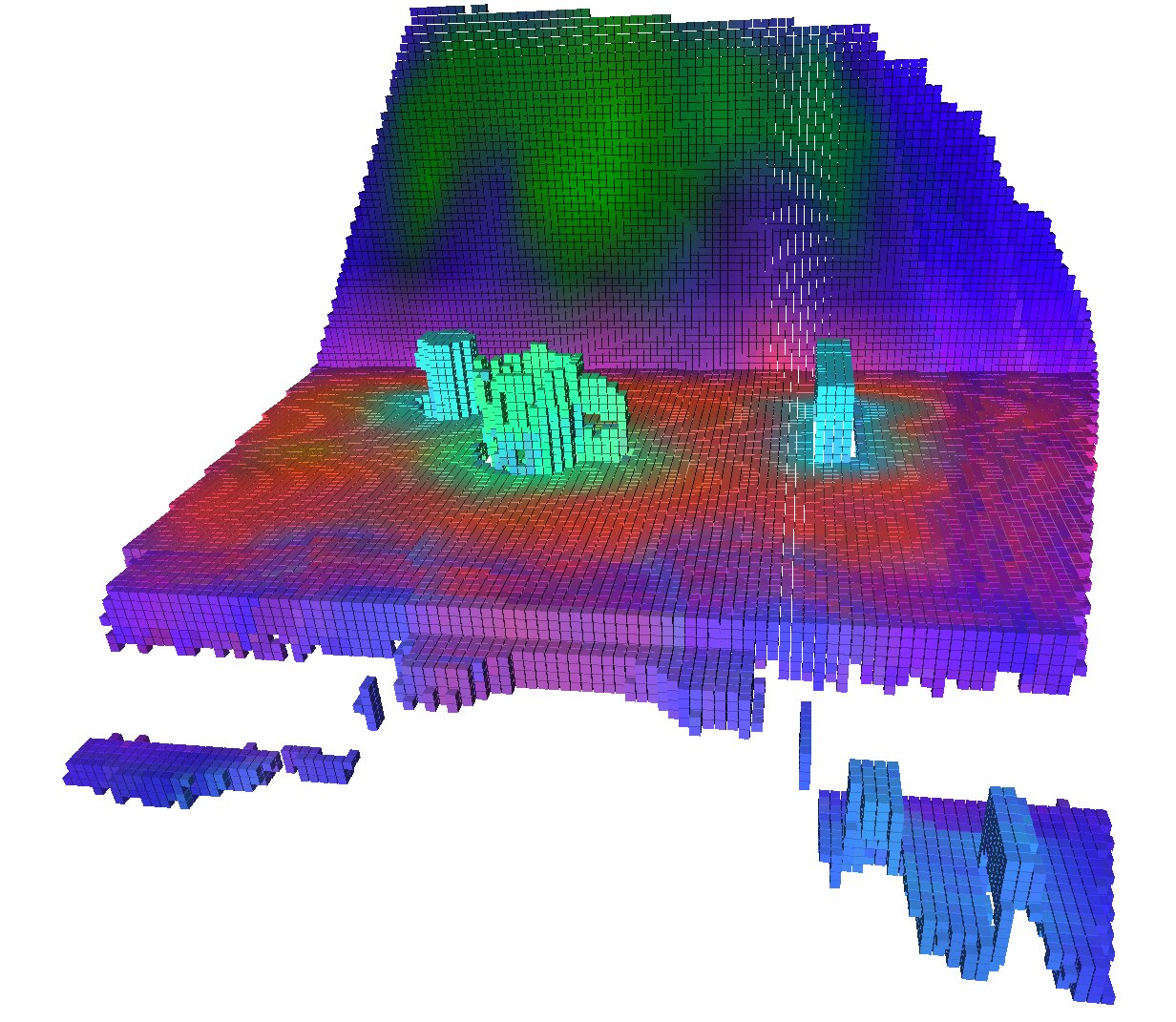}
    \\[5ex]  
    \includegraphics[width=0.40\linewidth]{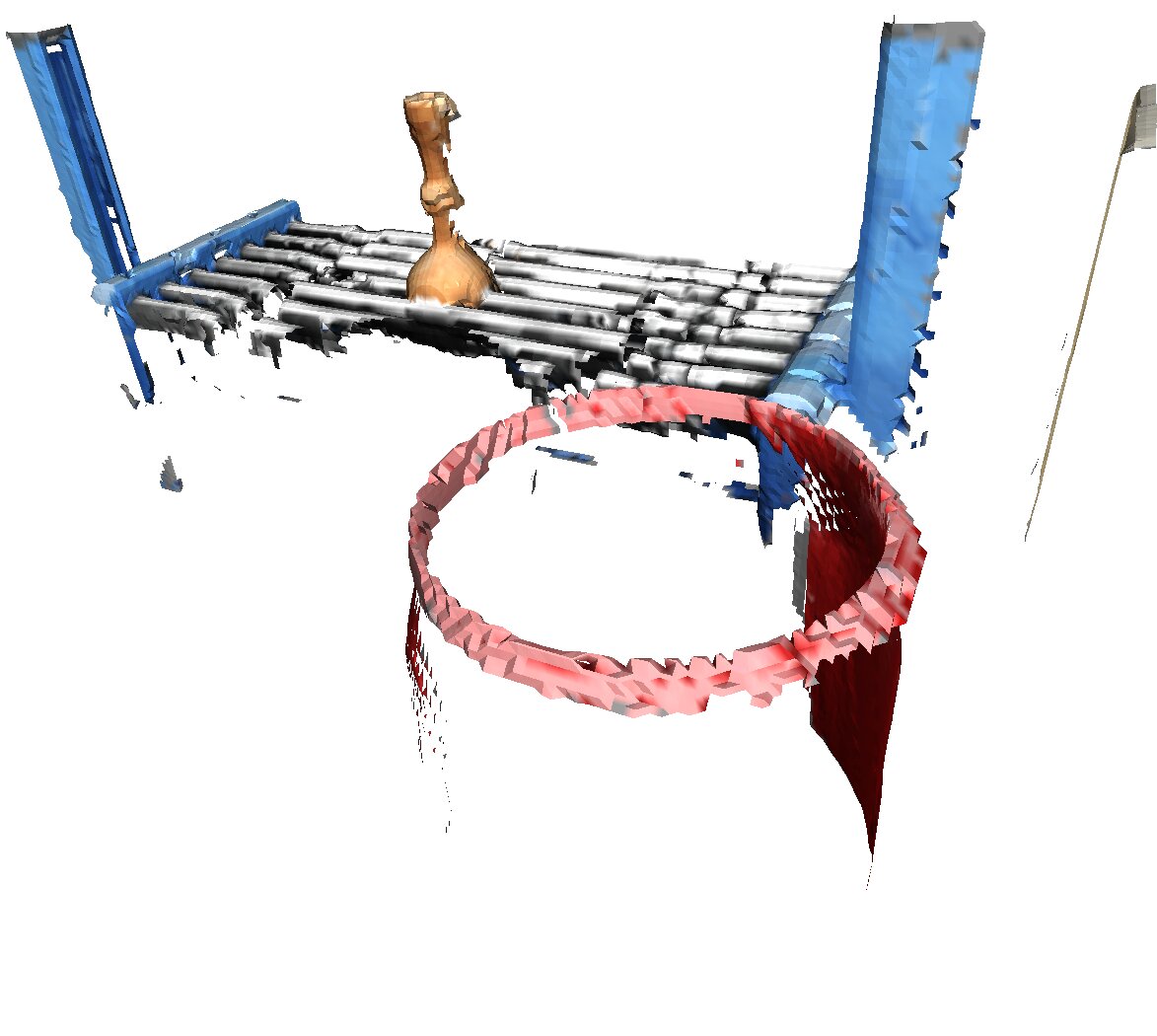}
    \hspace{2.0em}
    \includegraphics[width=0.40\linewidth]{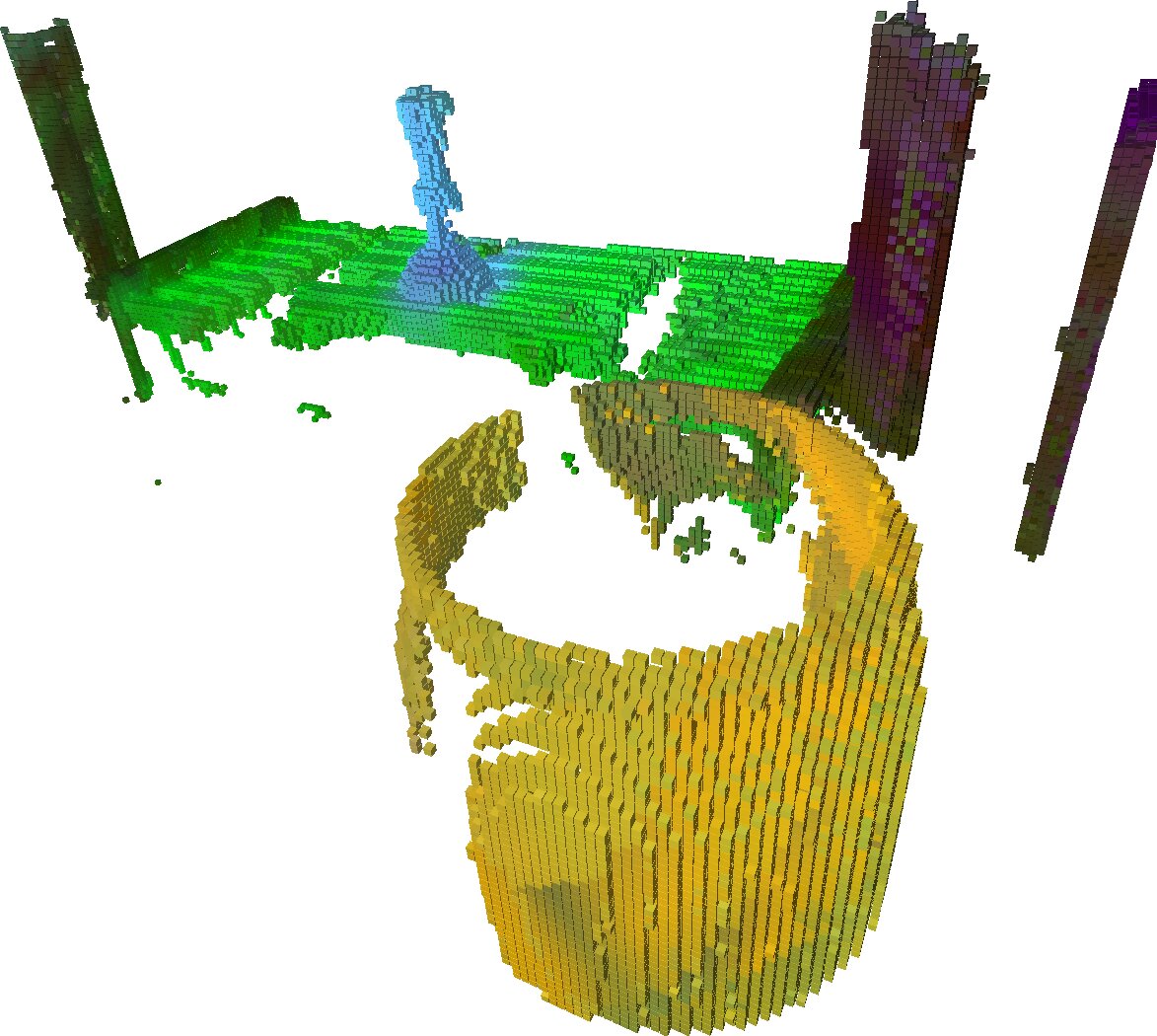}
    \\[5ex]  
    \includegraphics[width=0.40\linewidth]{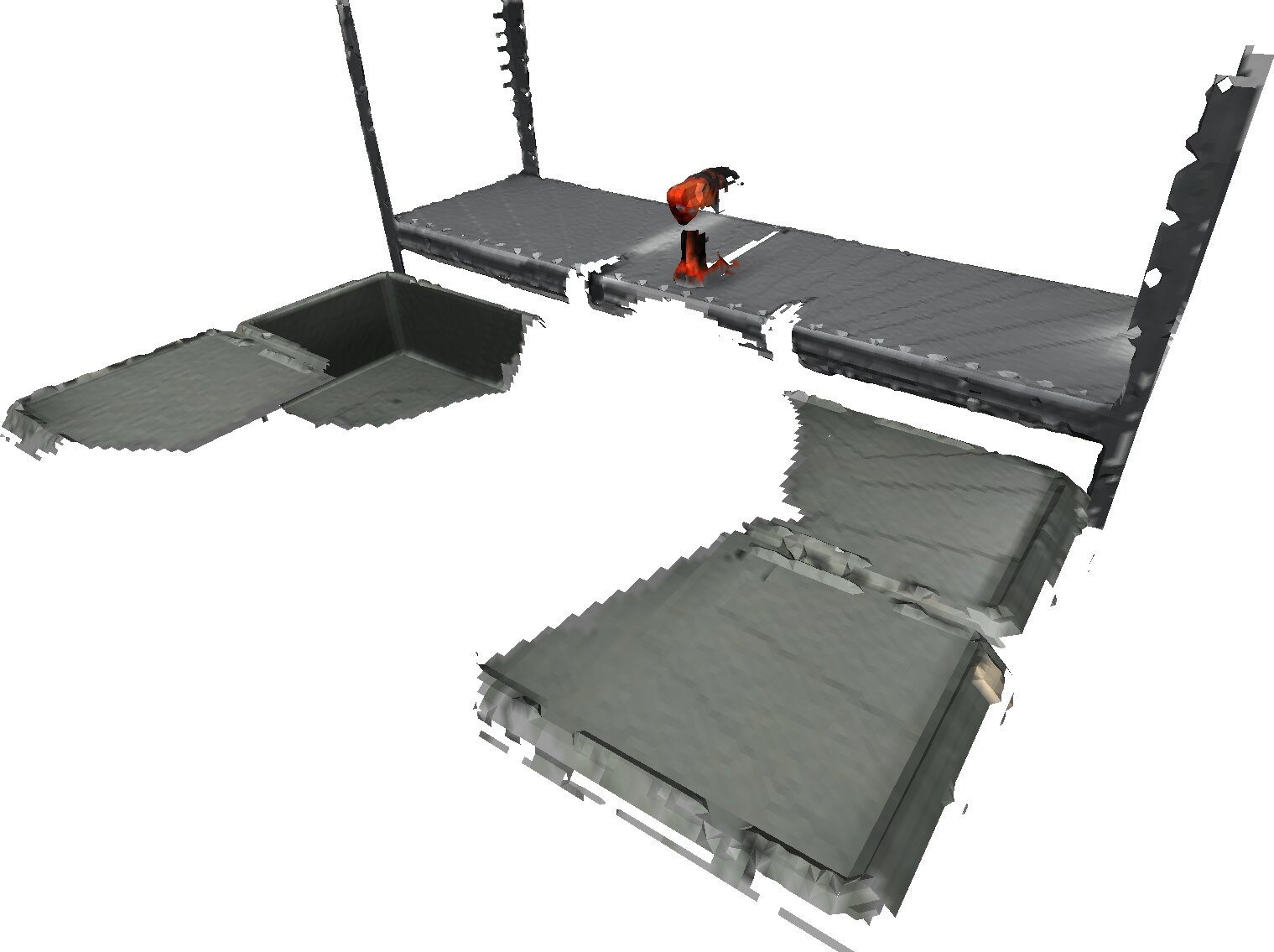}
    \hspace{2.0em}
    \includegraphics[width=0.40\linewidth]{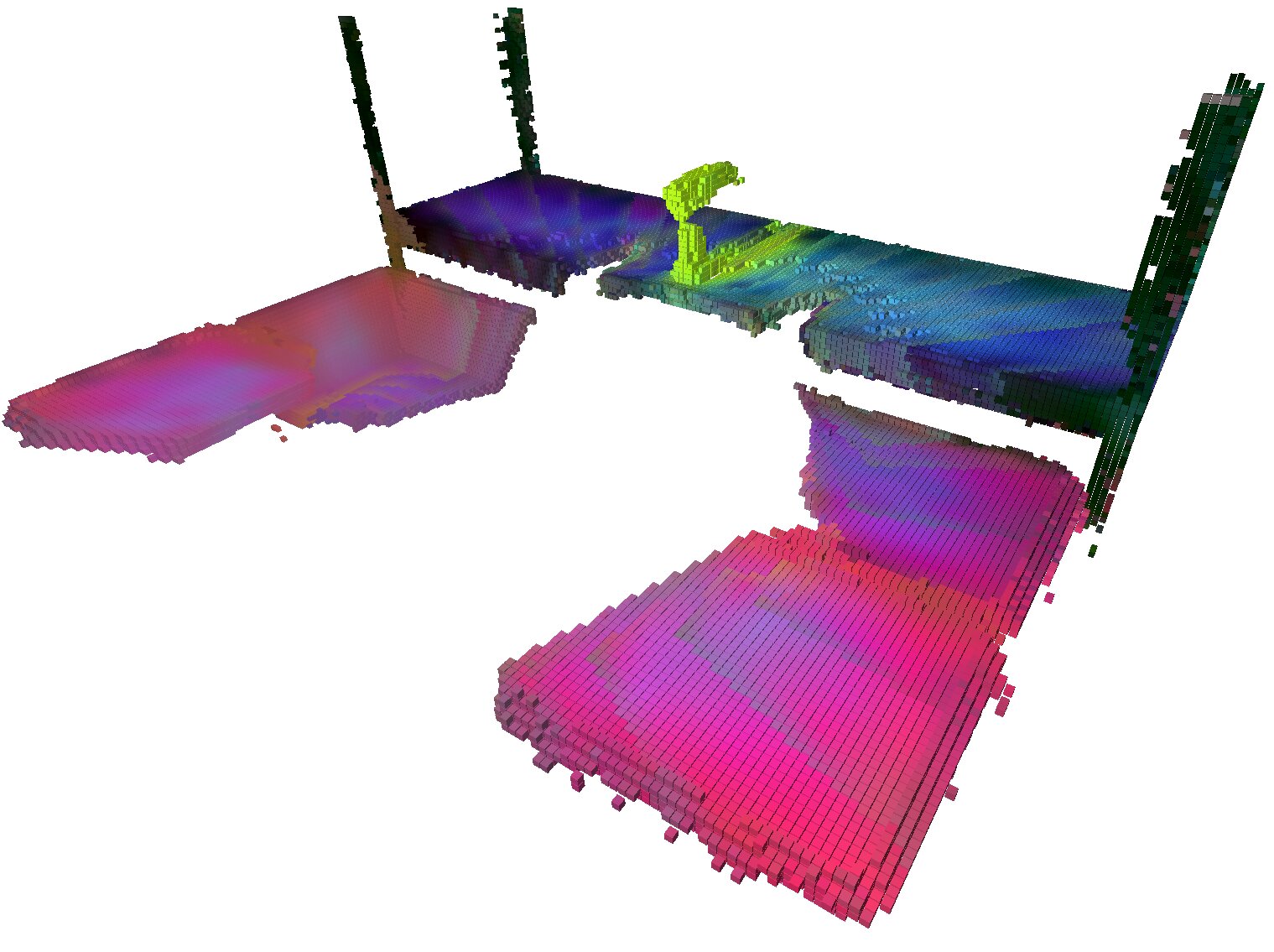}
    \\[5ex]  
    \caption{Reconstructions of the four environments presented in \refsec{sec:result}. For each environment, we have an RGB-colored mesh (top) and the voxel grid containing \ac{VFM} features colored by \ac{PCA} (bottom).}
    \label{fig:reconstructions}
\end{figure}

Reconstructions of our environments can be seen in \reffig{fig:reconstructions}.

\subsection{Evaluation task descriptions}
\label{appendix:task_descriptions}

\begin{figure}
    \centering
    \includegraphics[width=0.32\linewidth]{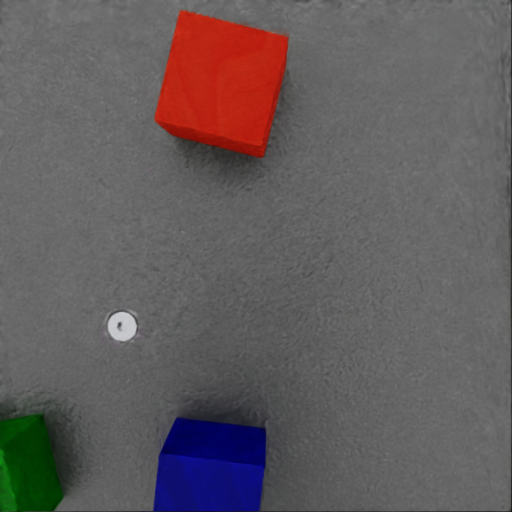}
    \hspace{0.1em}
    \includegraphics[width=0.32\linewidth]{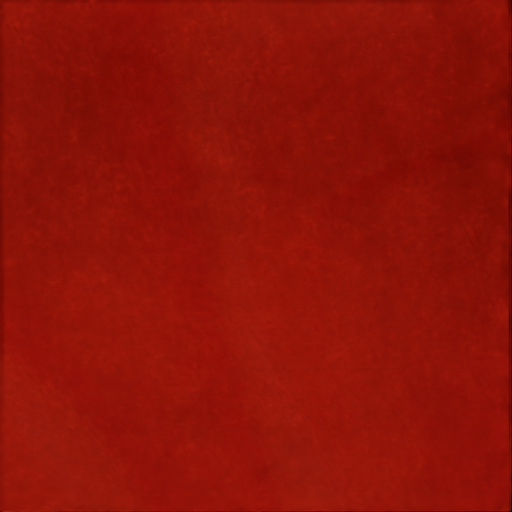}
    \hspace{0.1em}
    \includegraphics[width=0.32\linewidth]{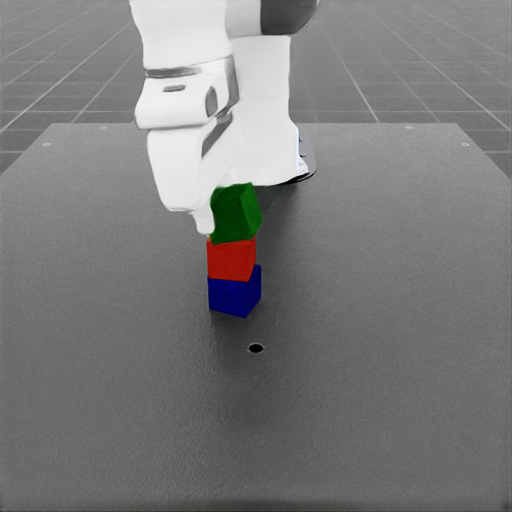}
    \\[1ex]  
    \includegraphics[width=0.32\linewidth]{images/teaser/pov_2_compressed.jpg}
    \hspace{0.1em}
    \includegraphics[width=0.32\linewidth]{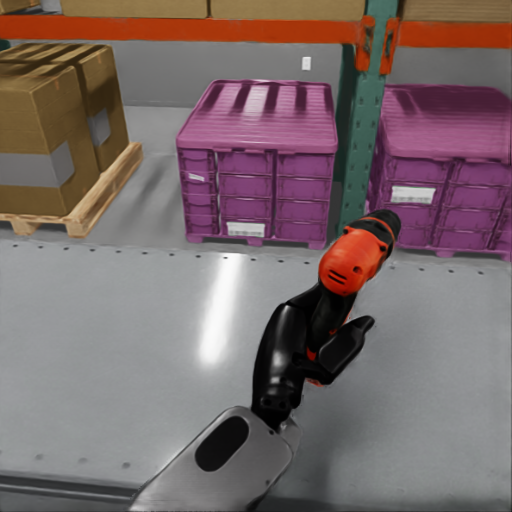}
    \hspace{0.1em}
    \includegraphics[width=0.32\linewidth]{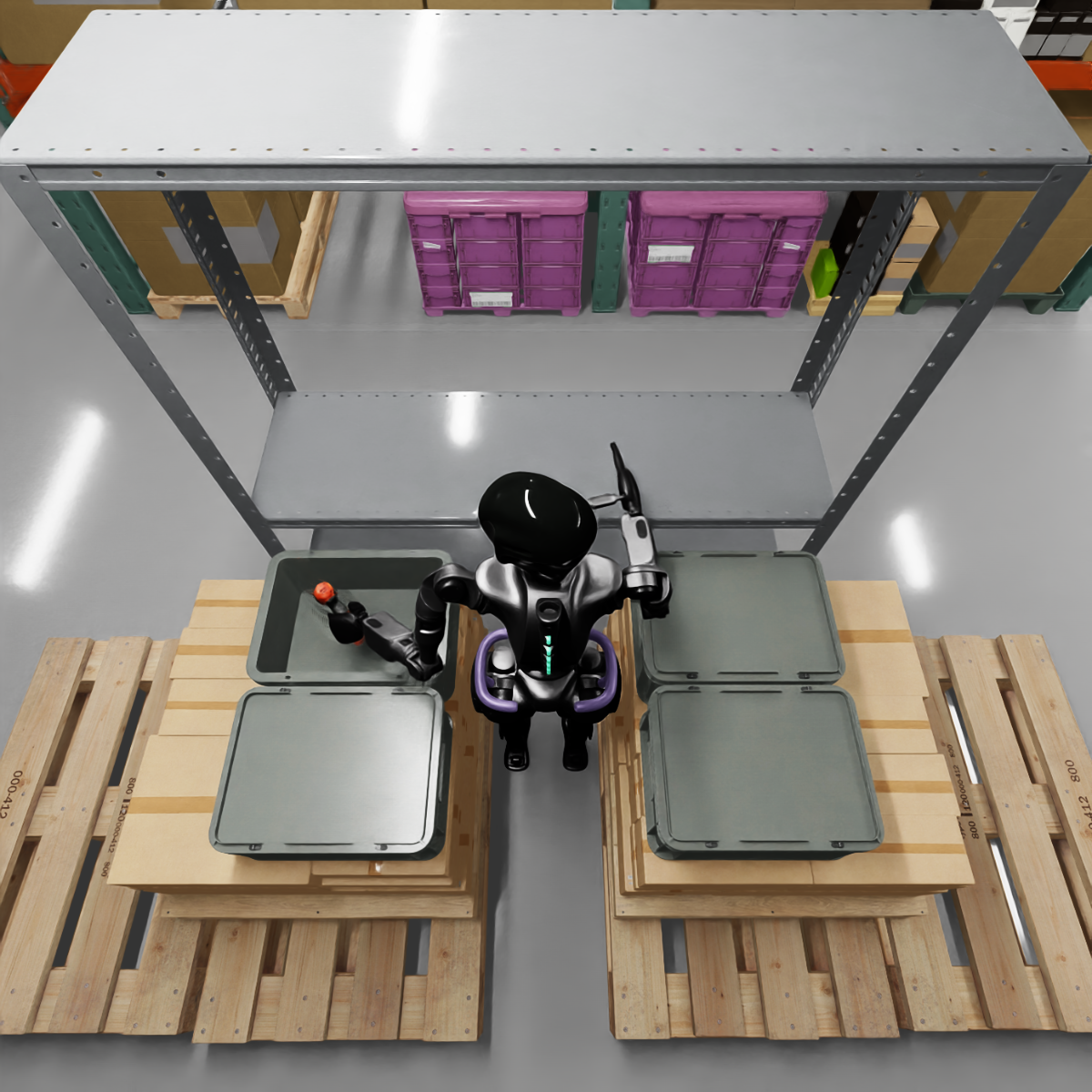}
    \\[1ex]  
    \includegraphics[width=0.32\linewidth]{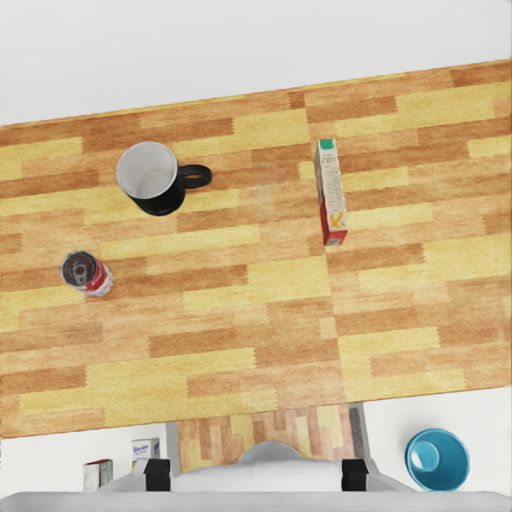}
    \hspace{0.1em}
    \includegraphics[width=0.32\linewidth]{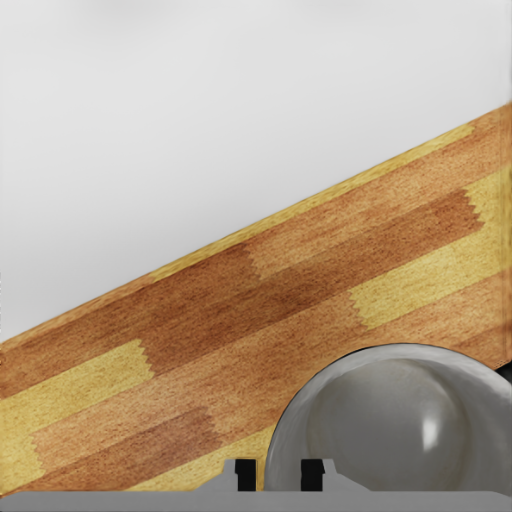}
    \hspace{0.1em}
    \includegraphics[width=0.32\linewidth]{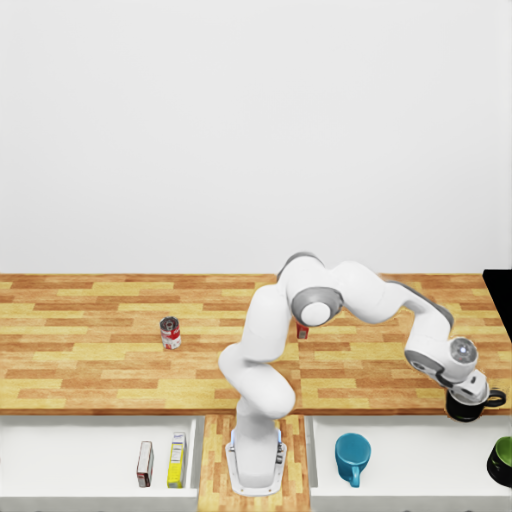}
    \\[1ex]  
    \includegraphics[width=0.32\linewidth]{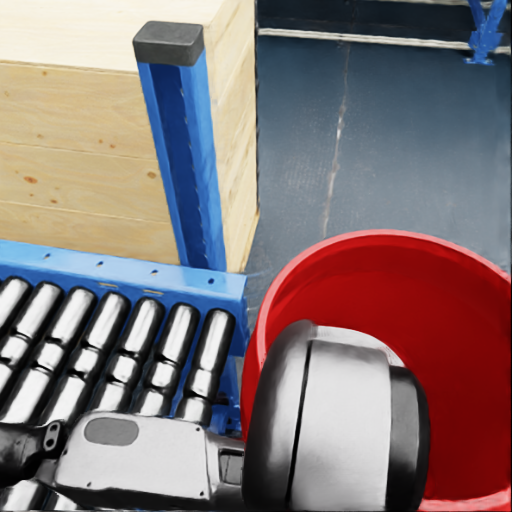}
    \hspace{0.1em}
    \includegraphics[width=0.32\linewidth]{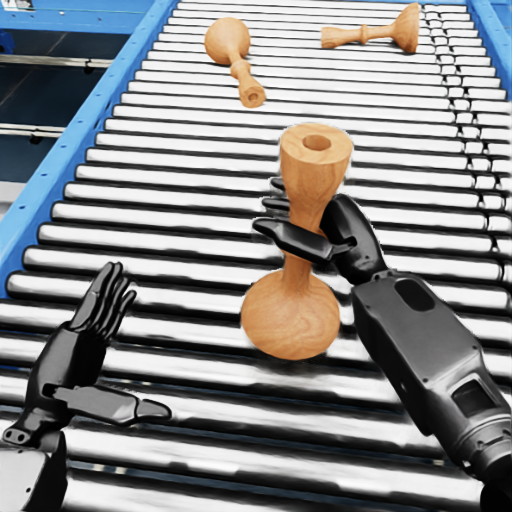}
    \hspace{0.1em}
    \includegraphics[width=0.32\linewidth]{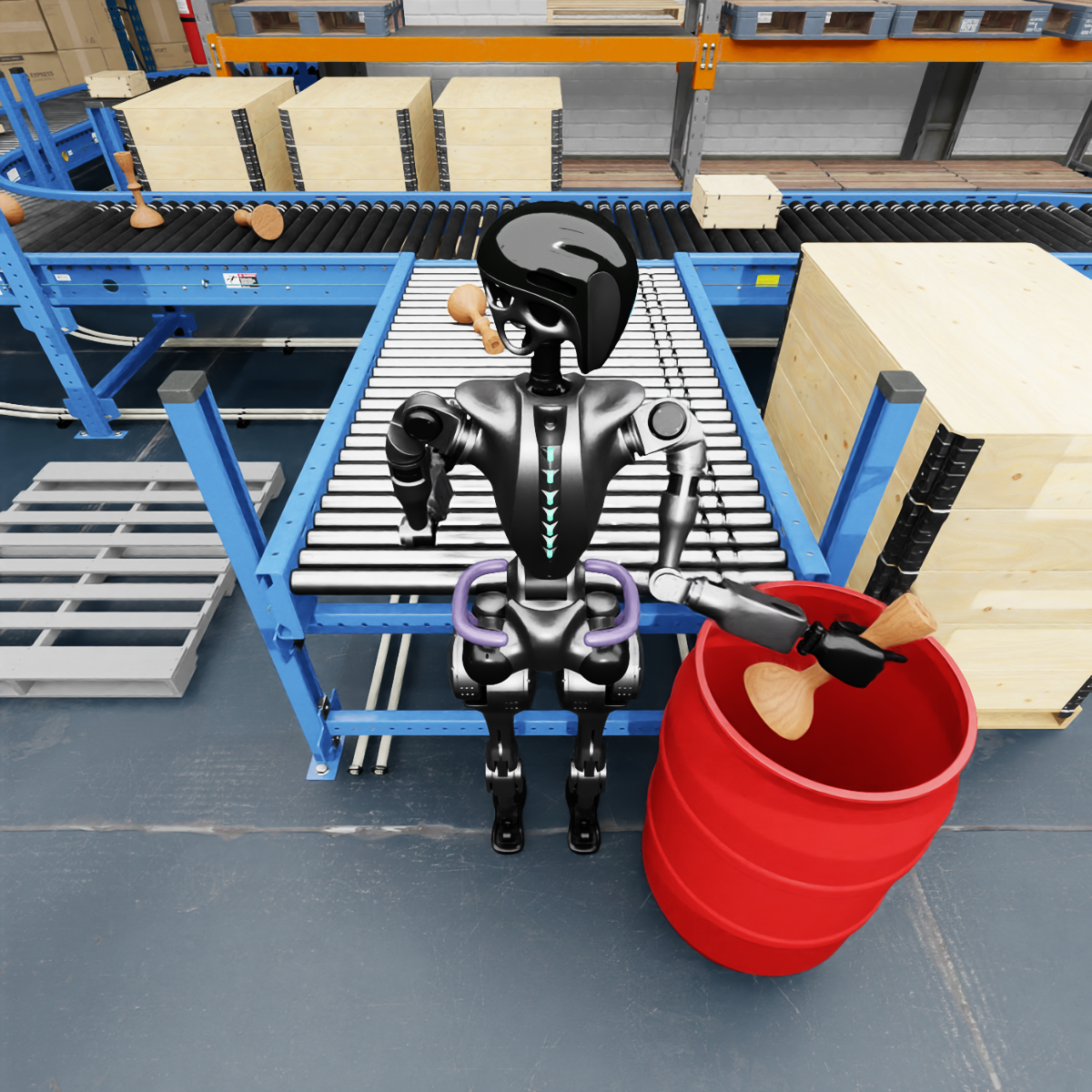}
    \caption{Views from four evaluation environments. Each row corresponds to a distinct environment. The first column presents an ego-centric perspective of the drop off locations, whose positions must be memorized. The second column shows ego-centric observations during task execution, where parts of these objects are no longer visible. The third column presents a third-person view of the robot performing the task.
    }
    \label{fig:environment_views}
\end{figure}

We introduce several tasks specifically designed to test for systems for their ability to leverage spatial memory.
See \reffig{fig:environment_views} for visualizations of the tasks.
In particular, we introduce:

\begin{itemize}[itemsep=0.5ex, parsep=0pt]
    \item \textbf{\cubestacking (robot arm):} Requires the policy to stack three cubes in order. Cube positions are randomized. The policy only has an egocentric view, and as a result, the policy must remember the position of the ongoing stack during cube transport, during which time the camera is blocked.
    \item \textbf{\mugindrawer (robot arm):} The goal of the task is to return a mug to a drawer that contains mugs. The target drawer is permuted between two options. The policy only has an egocentric view, and as a result, the policy must remember which of the two drawers is correct during transport of the mug.
    \item \textbf{\drillinbox (humanoid):} This task requires the humanoid robot to pick up an electric drill off the shelf and place it in an open box. Which box is open is randomly permuted among four options. To identify which is the correct box, the humanoid must actively scan its surroundings by rotating its head to detect the open box, memorize its location, and subsequently transport the drill to that position.
    \item \textbf{\stickinbin (humanoid):} 
    Similar to above. The humanoid robot must place a candlestick in a bin. The bin is randomly placed in a position around the robot. Successful task completion requires first scanning the scene, memorizing the layout, before transporting the stick.
\end{itemize}

\subsection{Reconstructing with nvblox-PyTorch}
\label{appendix:nvblox_torch}

\nvbloxtorch \cite{millane2024} is an open source library for real-time 3D reconstruction, designed for robotic applications. It provides functions for building, manipulating and querying 3D reconstructions directly on the GPU. The following snippet demonstrates how \mindmap makes use of the recently added PyTorch bindings to generate a featurized 3D reconstruction.

\begin{minted}[fontsize=\small, linenos]{python}
# Install nvblox_torch from pip
from nvblox_torch import Mapper, FeatureMesh

# Create a mapper.
mapper = Mapper(voxel_sizes_m=[0.01])

# Add depth and feature frames to the reconstruction.
for depth_frame, feature_frame, pose, intrinsics in dataset: 
    mapper.add_depth_frame(depth_frame, pose, intrinsics)
    mapper.add_feature_frame(feature_frame, pose, intrinsics)

# Compute a surface mesh representation of the scene.
mapper.update_feature_mesh()
mesh = mapper.get_feature_mesh()

# Obtain features and vertices as PyTorch tensors.
vertices = mesh.vertices()
features = mesh.features()
\end{minted}

\end{document}